\pdfoutput=1

\documentclass[11pt]{article}

\usepackage[]{acl}
\usepackage{times}
\usepackage{latexsym}
\usepackage{multirow}
\usepackage{graphicx}
\usepackage{amsmath}

\usepackage{colortbl}
\usepackage{booktabs}

\usepackage[capitalize]{cleveref}

\usepackage{svg}
\usepackage{bbding}

\usepackage{pifont}
\usepackage{amssymb}

\usepackage{graphicx}
\usepackage{adjustbox}

\usepackage[T1]{fontenc}

\usepackage{booktabs}

\usepackage[utf8]{inputenc}

\usepackage{microtype}

\usepackage{inconsolata}
\usepackage{xspace}

\newcommand{\logofig}{\includegraphics[height=0.78em,trim=0 0.5em 0 0]{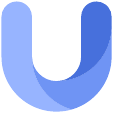}}

\newcommand{\logott}{\logofig\textsc{Bench}\xspace}

\title{\logott: Benchmarking Uncertainty in Large Language Models with Multiple Choice Questions}

\author{Xunzhi Wang$^{1, 4}$\thanks{Equal contribution.} \quad Zhuowei Zhang$^{1, 4}$\footnotemark[1] \quad Gaonan Chen$^1$ \quad Qiongyu Li$^1$  \\
\textbf{\quad Bitong Luo$^1$ \quad Zhixin Han$^1$ \quad Haotian Wang$^1$}\\
\textbf{\quad Zhiyu li$^2$ \quad Hang Gao$^3$ \quad Mengting Hu$^{1, 4}$\thanks{Corresponding author.}}\\ 
$^1$College of Software, Nankai University 
\quad $^2$Institute for Advanced Algorithms Research (Shanghai) \\ $^3$College of Artificial Intelligence, Tianjin University of Science and Technology \\ 
$^4$Tianjin Key Laboratory of Software Experience and Human Computer Interaction  \\
{xunzhi, zhuoweizhang}@mail.nankai.edu.cn \quad {lizy@iaar.ac.cn} \quad {mthu@nankai.edu.cn}}

\begin{document}
\maketitle
\begin{abstract}

Despite recent progress in systematic evaluation frameworks, benchmarking the uncertainty of large language models (LLMs) remains a highly challenging task. 
Existing methods for benchmarking the uncertainty of LLMs face three key challenges: the need for internal model access, additional training, or high computational costs. This is particularly unfavorable for closed-source models. To this end, we introduce \textbf{\textsc{UBench}}, a new benchmark for evaluating the uncertainty of LLMs. Unlike other benchmarks, \textbf{\textsc{UBench}} is based on confidence intervals. It encompasses 11,978 multiple-choice questions spanning knowledge, language, understanding, and reasoning capabilities. Based on this, we conduct extensive experiments. This includes comparisons with other advanced uncertainty estimation methods, the assessment of the uncertainty of 20 LLMs, and an exploration of the effects of Chain-of-Thought (CoT) prompts, role-playing (RP) prompts, and temperature on model uncertainty. Our analysis reveals several crucial insights: \textit{1) Our confidence interval-based methods are highly effective for uncertainty quantification; 2) Regarding uncertainty, outstanding open-source models show competitive performance versus closed-source models; 3) CoT and RP prompts present potential ways to improve model reliability, while the influence of temperature changes follows no universal rule.} Our implementation is available at https://github.com/Cyno2232/UBENCH.  
\end{abstract}

\section{Introduction}
In recent years, significant progress has been made in the development of large language models (LLMs), including ChatGPT \cite{ChatGPT}, Llama \cite{Llama3}, Qwen \cite{qwen2,qwen2.5}, etc.
These advancements have not only piqued the intense interest of both the industrial and academic sectors \cite{zhao2023survey}, but have also spurred a wave of exploration and research. In this fast-paced development, comprehensively evaluating LLMs' capabilities is essential. This evaluation helps us understand LLMs better, improve their performance, and apply them effectively in real-world scenarios, unlocking their potential for innovation across different fields.

\begin{figure}[!t]
    \centering
    \includegraphics[width=0.9\linewidth]{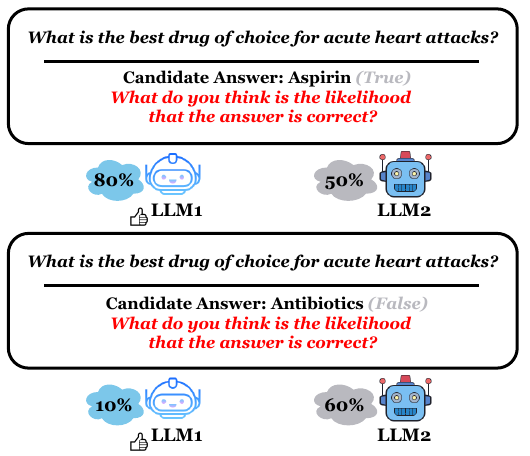}
  \caption{
   In the context of different candidate answers to the same question, LLMs display different levels of confidence (in other words, uncertainty). Note that LLMs may exhibit consistent levels of confidence for either the wrong answer or the right answer, which we do not want.}
  \label{fig:intro}
\end{figure}

Much effort has been devoted to establishing accurate, authentic, and equitable evaluation systems for LLMs, such as C-Eval \cite{C-Eval} and MT-Bench \cite{MTbench}. However, these benchmarks often focus solely on the performance of models across various tasks, neglecting the credibility of their answers. In some situations, models may not be certain about the confidence level associated with their provided answers, potentially leading to misunderstandings or even harm. Specifically, LLMs are prone to generating misinformation without warning, which may manifest as hallucinations \cite{huang2023survey}, biases \cite{felkner2023winoqueer}, or disinformation \cite{lucas2023fighting}. In this scenario, we hope to know the confidence level of the response to decide whether to trust the information or suggestions provided by LLMs. Specifically, as shown in Figure \ref{fig:intro}, two LLMs exhibit different confidence levels for various candidate answers to the same question \emph{``What do you think is the likelihood that the answer is correct?''}. Clearly, the performance of LLM1 aligns more closely with human expectations. Inspired by that, some research has initiated a shift towards assessing the uncertainty of LLMs \cite{iclr,tengxun,xu-etal-2024-sayself}. 

However, many existing uncertainty estimation methods for LLMs are severely restricted. They confront challenges like the need for additional training, high computational costs, and access to internal model details. This is especially tough for closed-source models. These limitations hamper the application of uncertainty estimation, and thus, a comprehensive and accurate assessment of model reliability and performance. 
For instance, \citet{xu-etal-2024-sayself} proposed SaySelf, a two-stage training framework that requires extra training, but it is difficult to be used for closed-source models. \citet{iclr} use prompts to get model confidence, but it needs multiple sampling, which is costly and not ideal in some cases\footnote{Please refer to \S\ref{Case Study} for details.}. \citet{tengxun}'s logits-based method for uncertainty assessment needs internal model access, so it only works for open-source models. Approximating it for closed-source models via multiple sampling, like \citet{iclr}'s method, is inefficient and resource-consuming.

\begin{table}[t]
\centering

\small

\begin{tabular}{lccccc} 
\toprule
\textbf{Method} & \textbf{Close?} & \textbf{Single?} &\textbf{Close\&Single?}\\ 
\midrule  
\citet{iclr} & \checkmark & \ding{55} & \ding{55} \\
\citet{tengxun} & \checkmark  & \checkmark & \ding{55} \\

\textsc{UBench} (Ours) & \checkmark & \checkmark & \checkmark  \\
\bottomrule
\end{tabular}
\caption{Comparison of different training-free uncertainty estimation methods: \textbf{\textit{“Close?”}} indicates whether the method can be applied to closed-source models, \textbf{\textit{“Single?”}} indicates whether only a single inference is needed, and \textbf{\textit{“Close\&Single?”}} indicates whether the method can be applied to closed-source models with single-inference requirements.
}
\label{Comparison}
\end{table}
To address these challenges, we introduce \textbf{\textsc{UBench}}, a new benchmark. Distinct from previous efforts, it is founded on confidence intervals, which not only facilitates automated evaluation but also enables more effective quantification of uncertainties. Comprising four question categories, \textsc{UBench} encompasses a total of 11,978 (\textasciitilde 12K) multiple-choice questions, including both positive and negative samples. Notably, to improve efficiency while maintaining accuracy, some of the negative samples were generated by GPT-4\footnote{The version is 1106-preview.} and subsequently underwent manual review and refinement (refer to \S\ref{Data Validation Details} for details). Designed to accurately assess the reliability of LLMs, \textsc{UBench} is resource-efficient, requiring no extra training, only one inference, and no internal model parameters, filling a key gap in LLM evaluation. A comparison of \textsc{UBench} with other training-free benchmark uncertainty methods is presented in Table \ref{Comparison}\footnote{Please refer to \S\ref{Results of UBench compared to other methods} for more results.}.

Based on this, we compare the uncertainty quantification method based on confidence intervals with other powerful methods. We also evaluate 20 state-of-the-art open-source and closed-source models that cover several series and sizes. In addition, we explore the factors that potentially influence model uncertainty, including Chain-of-Thought (COT) \cite{wei2022chain} prompts, role-playing (RP) \cite{ shao2023character} prompts, and the impact of temperature on model uncertainty. Our main contributions and findings are summarised below:

\begin{itemize}
    \item We propose \textsc{UBench}, a new confidence interval-based uncertainty evaluation benchmark for LLMs. It consists of approximately 12K questions, covering four categories: knowledge, language, understanding, and reasoning.
    \item We conduct a comparison of our confidence interval-based uncertainty estimation method with other LLM uncertainty estimation methods and achieve superior results.
    \item We utilize \textsc{UBench} to conduct tests on 20 widely-adopted LLMs. In general, excellent open-source and closed-source LLMs display comparable degrees of reliability. 
    \item Further exploratory analysis reveals that CoT and RP prompts are potential methods for enhancing model reliability, while there is no general rule regarding the impact of temperature. We analyze the raw responses of LLMs and offer possible explanations for each effect. This analysis paves the way for broader downstream applications of LLMs. 
\end{itemize}

\begin{figure*}[!t]
  \centering
\includegraphics[width=0.985\linewidth]{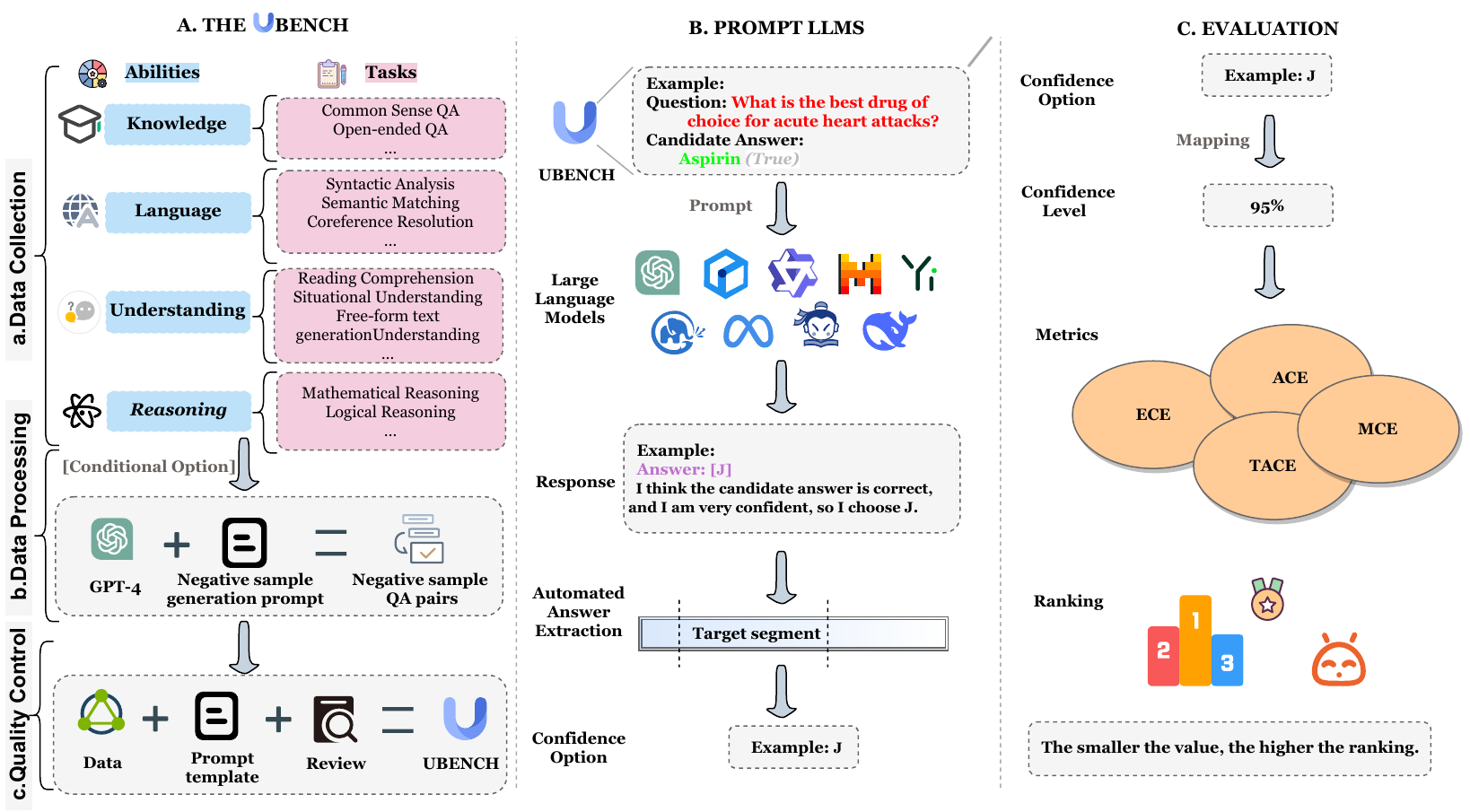}
  \caption{
    Construction process of \textsc{UBench} and systematic, automated LLM uncertainty evaluation framework. The data sources of \textsc{UBench} are from multiple types of public datasets, with processed into the uncertainty evaluation format and carefully controlled quality. Then \textsc{UBench} is leveraged to compare the reliability of typical open-source and closed-source LLMs with 4 evaluation metrics.
  }
  \label{fig:framework}
\end{figure*}

\section{Related Work}
\subsection{Benchmark for LLMs}

Previous benchmarks can generally be categorized into two types: generic task benchmarks and task-specific benchmarks \cite{chang2023survey}. Generic task benchmarks are used to evaluate the generic capabilities of LLMs for several tasks (e.g., sentiment analysis, natural language inference, machine translation, etc.), including GLUE \cite{Glue}, MMLU \cite{MMLU}, MT-Bench \cite{MTbench}, BIG-bench \cite{BIG-bench}, PromptBench \cite{PromptBench}, PandaLM \cite{PandaLM}, MT-Eval \cite{kwan2024mt} and so on. \citet{bai-etal-2024-compound} present ChatABSA, specifically designed to assess LLMs' performance on aspect-based sentiment analysis (ABSA). ToMBench \cite{chen2024tombench} is used to benchmark the Theory of Mind in large language models.
C-Eval \cite{C-Eval} is the first benchmark for broadly assessing a model's Chinese knowledge and reasoning ability.

\subsection{Uncertainty Estimation for LLMs} \label{Uncertainty Estimation for LLMs}

Recent research in uncertainty estimation for LLMs has explored diverse methodologies, broadly categorized into logits-based, verbal-based, and training-based approaches. Logits-based methods leverage model output and parameters to assess uncertainty, though their applicability is constrained. For instance, \citet{kuhn2022semantic} cluster semantically similar answers to evaluate uncertainty, while \citet{duan2023shifting} incorporate sentence relevance for uncertainty estimation; however, these methods are unsuitable for closed-source LLMs. Although \citet{tengxun}'s method can approximate results by repeatedly sampling the closed-source model, it exacerbates resource consumption.
Verbal-based confidence assessment methods require models to output confidence scores in natural language, as proposed by \citet{lin2022teaching}. While \citet{tian-etal-2023-just} observe better calibration in RLHF-LMs’ confidence scores, \citet{iclr} highlight overconfidence issues and advocate combining verbalized confidence with consistency. Furthermore, the applicability of these two evaluation methods is still limited to fact-recall tasks or multi-step inference requirements. Training-based methods often require retraining or fine-tuning, limiting scalability. Examples include UaIT \citep{liu-etal-2024-llms-learn-uncertainty}, which aligns uncertainty perception with probabilistic outputs, and R-Tuning \citep{zhang-etal-2024-r} for enhancing refusal capability on uncertain queries. Moreover, \citet{bakman-etal-2024-mars} propose MARS that takes into account the semantic contribution of each token to evaluate the correctness of the generated content. \citet{xu-etal-2024-sayself} mitigate this with Sayself, a reinforcement learning framework for calibrated confidence generation. For more analysis, please refer to~\S\ref{More Related Work}.

Unlike all the aforementioned works, our proposed \textsc{UBench} introduces a novel confidence interval-based sampling approach that achieves three key advantages: requiring only single inference, without requiring additional model training, and applicability to closed-source models. Furthermore, while most existing methods have only been validated on a limited number of tasks, \textsc{UBench} serves as a multi-dimensional, multi-task benchmark designed to comprehensively evaluate the uncertainty of LLMs.

\section{The \logott}
The overall construction and evaluation process of \textsc{UBench} is shown in Figure \ref{fig:framework}. Overall, \textsc{UBench} includes 4 categories, including tasks such as reading comprehension, QA, mathematical reasoning, and more. We provide detailed information on categories, data construction, and the design of prompts.

\subsection{Problem Categories}
We adopt the competency categorization from OpenCompass's LLM assessment framework \cite{2023opencompass}, resulting in the following four categories:

\noindent
\textbf{\includegraphics[width=0.5cm]{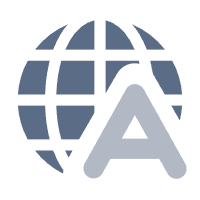} Language.} This category primarily evaluates the reliability of LLMs in dealing with language category tasks, encompassing tasks such as syntactic analysis, semantic matching, word sense disambiguation, and coreference resolution.

\noindent
\textbf{\includegraphics[width=0.5cm]{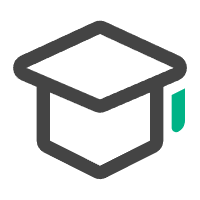} Knowledge.} This type of question mainly evaluates the reliability of LLMs when dealing with knowledge-based tasks such as common sense and facts, covering contents in areas like health, law, finance, politics, and history.

\noindent
\textbf{\includegraphics[width=0.5cm]{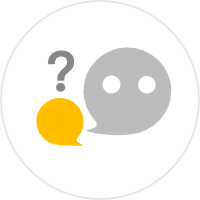} Understanding.} The aim is to evaluate LLMs' reliability in understanding-related tasks. These tasks encompass various aspects, including multitask language understanding, reading comprehension, situational awareness, behavioral interpretation, dialogue understanding, and so on.

\noindent
\textbf{\includegraphics[width=0.5cm]{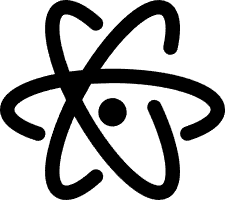} Reasoning.} Unlike other categories, this one aims to evaluate LLMs' reliability in mathematical reasoning, logical reasoning, and related areas.

\subsection{Data Construction}
The data sources of \textsc{UBench} originate from various existing datasets and are specially processed into formats suitable for evaluation, with careful quality control.

\vspace{2pt}
\noindent
\textbf{Data Collection.} \; To comprehensively and efficiently construct the dataset, we conduct random sampling from 24 open-source datasets\footnote{Most of the datasets used in this study are sourced from \href{https://huggingface.co/datasets}{HuggingFace}.}. Each dataset contributes 500 data points, as shown in Figure \ref{fig:benchmark}. The datasets sampled for each category are listed in the Appendix \ref{Datasets Overview}.

\vspace{2pt}
\noindent
\textbf{Data Processing.} \; Intuitively, a reliable LLM yields lower uncertainty for correct answers and higher uncertainty for incorrect ones. Therefore, different from previous works, we reformat the collected data into positive and negative samples, respectively. A positive sample indicates that the correct answer is used, and a negative sample indicates that an incorrect answer is randomly selected as the answer. For datasets without candidate wrong answers, we prompt GPT-4-1106-preview using a one-shot approach to generate incorrect answers similar to the correct ones. For details on the prompts, please refer to Appendix~\S\ref{Prompt Templates}.

\vspace{2pt}
\noindent
\textbf{Quality Control.} \;
On one hand, for the generated negative samples, we conduct a comprehensive check of all data. Failed data are regenerated until they meet the requirements, ensuring data integrity for subsequent analysis.
On the other hand, to ensure dataset quality, a strict review process is implemented. Each sample is first reviewed by two authors. Only when both concur that there are no issues is the sample approved. In case of disagreement, a third author joins the review to reach a consensus. Through this process, the sample approval rate has reached 99.82\%. For more details on the data validation process, refer to Appendix~\S\ref{Data Validation Details}.

\begin{figure}[t!]
  \centering
  \includegraphics[width=\linewidth]{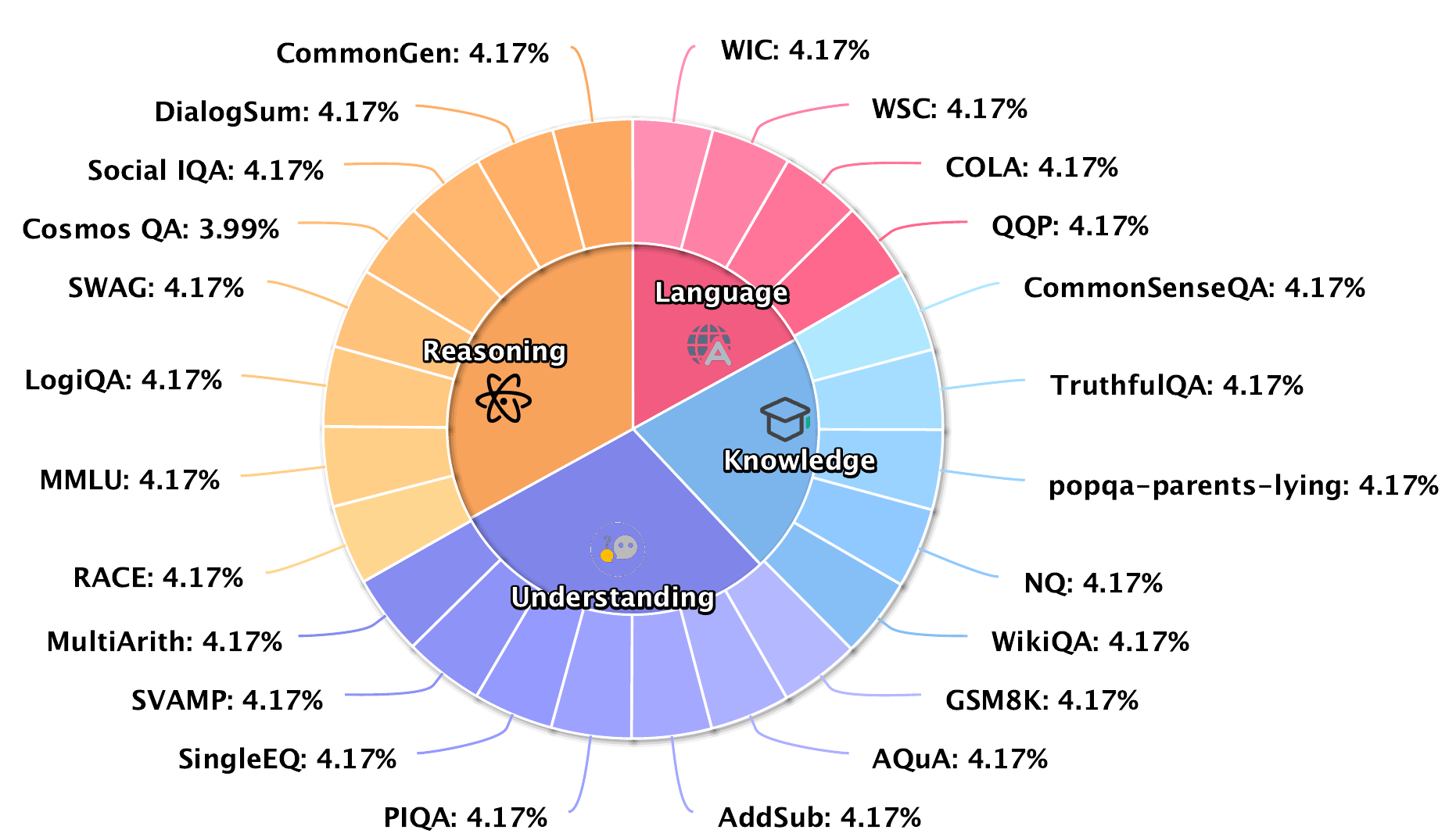}
  \caption{
    \textsc{UBench} covers 4 categories: Language, Knowledge,  Understanding, and Reasoning, with a total of 11,978 ten-item multiple-choice questions.
  }
  \label{fig:benchmark}
\end{figure}

\subsection{Prompt Design}

Like previous studies \cite{zhang2023safetybench, MTbench}, we evaluate the reliability of LLMs based on prompt engineering. Initially, we design the prompt with a data sample and all its answers, following instructions like \emph{choose the correct answer and also yield the uncertainty.} We call this the standard prompt. However, most LLMs struggle to output a consistent format of uncertainty values, even some of them do not understand the instructions~\cite{kuhn2023semantic}, which makes it hard to evaluate. Therefore, we formulate the prompt as providing a correct answer or incorrect answer to LLMs, asking them to choose the uncertainty towards the given answer from ten intervals. In other words, the problem is reformulated from yielding real number uncertainty to a multiple-choice question. We refer to this as the confidence interval-based prompt. For empirical validation on prompt design, please refer to~\S\ref{Empirical validation on prompt design}.

Specifically, we create 10 confidence intervals, each representing a 10\% confidence range, e.g. 0-10\% for the first interval, 10-20\% for the second, and so on. Different annotators having varying criteria for confidence, results in annotation errors, making it difficult to construct few-shot prompts. We choose a zero-shot approach to construct the prompt instead. The prompts contain role-playing (RP) prompts, task declarations, and a step-by-step problem disassembly Chain-of-Thought (CoT) prompt, output format specification, confidence interpretation, sample QA, and confidence interval options. Please refer to Table~\ref{tab:prompt} for details.

\section{Experiments}

\subsection{Experimental Settings}

\textbf{Evaluated Models.} \; Based on \textsc{UBench}, we test 20 popular LLMs, covering a wide range of open and closed source LLMs, as shown in Table \ref{tab:LLMs}. 

\vspace{3pt}
\noindent
\textbf{Evaluation Metrics.}\label{Evaluation Metrics} \; To evaluate the reliability of LLMs, we use four evaluation metrics: Expected Calibration Error (ECE), which measures the difference between model confidence and accuracy \citep{Guo_Pleiss_Sun_Weinberger_2017}; Average Calibration Error (ACE), which adjusts for different confidence ranges; Maximum Calibration Error (MCE), indicating the worst-case confidence error \citep{Guo_Pleiss_Sun_Weinberger_2017}; and Thresholded Average Calibration Error (TACE) for high-risk scenarios, with a threshold set at 0.5.
More calculation details are shown in Appendix \S\ref{Details for evaluation Metrics}. All metrics are better with lower values and are presented as percentages in our work.

\begin{table*}[!t]
\centering
\scriptsize
\begin{tabular}{c|c|c|c|c|c|c}
\midrule
\multirow{2}{*}{\textbf{Type}} &\multirow{2}{*}{\textbf{LLMs}} & \multirow{2}{*}{\textbf{Method}} & \multicolumn{2}{c|}{\textbf{ECE (\%)} $\downarrow$} & \multicolumn{2}{c}{\textbf{MCE (\%)} $\downarrow$} \\ \cmidrule{4-7}
                     &              &           & Cosmos QA & SWAG  & Cosmos QA & SWAG \\ \midrule
\multirow{10}{*}{Open-source} 
&\multirow{5}{*}{Mistral-Instruct-7B}
                     & \textsc{UBench} (Ours)            & \textcolor{magenta}{26.24$\pm1.68$}        & \textcolor{blue}{30.26$\pm0.66$}     & \textcolor{blue}{68.33$\pm23.09$}        & 72.60$\pm11.81$    \\
                     && \citet{tengxun}            & \textcolor{blue}{31.68$\pm1.13$}        & 36.51$\pm8.57$     & 68.47$\pm14.11$       & \textcolor{magenta}{68.55$\pm5.18$}    \\
                     && \citet{iclr}  & 52.11$\pm6.94$        & 52.05$\pm11.02$      & 77.08$\pm5.30$      & 84.98$\pm3.90$ \\
                    && \citet{Guo_Pleiss_Sun_Weinberger_2017}            & 49.38$\pm9.41$        & 45.65$\pm5.45$     & \textcolor{magenta}{55.10$\pm10.29$}       & \textcolor{blue}{69.92$\pm17.09$}  \\
                    && \citet{gal2016dropoutbayesianapproximationrepresenting}            & 41.99$\pm1.60$        & \textcolor{magenta}{26.34$\pm3.32$}     & 88.33$\pm0.85$       & 75.35$\pm13.01$   
                      \\\cmidrule{2-7}
 &\multirow{5}{*}{Llama3-instruct-8B}
                     & \textsc{UBench} (Ours)                 & \textcolor{magenta}{7.5$\pm0.92$}        & \textcolor{magenta}{5.82$\pm1.78$}    & \textcolor{magenta}{34.17$\pm1.44$}    & \textcolor{blue}{62.22$\pm11.82$}   \\
                     && \citet{tengxun}         & 31.75$\pm1.20$        & \textcolor{blue}{32.88$\pm5.05$}     & 69.22$\pm12.98$        & 65.13$\pm5.63$    \\
                     && \citet{iclr}     & \textcolor{blue}{19.76$\pm4.25$}     & 44.72$\pm11.39$       & 69.07$\pm14.96$       & 75.86$\pm10.67$    \\
                     && \citet{Guo_Pleiss_Sun_Weinberger_2017}     & 36.75$\pm0.87$     & 62.37$\pm2.79$       & \textcolor{blue}{37.64$\pm0.60$}       & 63.70$\pm3.56$    \\
                     && \citet{gal2016dropoutbayesianapproximationrepresenting}     & 31.11$\pm3.93$     & 44.01$\pm2.00$       & 53.75$\pm5.30$       & \textcolor{magenta}{57.63$\pm5.26$}    \\
                     \midrule
\multirow{7}{*}{Closed-source} 
& \multirow{2}{*}{GPT-3.5}
                     & \textsc{UBench} (Ours)                & \textcolor{blue}{25.72$\pm0.36$}       & \textcolor{magenta}{18.78$\pm1.58$}     & \textcolor{blue}{50.26$\pm17.60$}    & \textcolor{blue}{78.33$\pm5.77$}    \\
                      &&  \citet{iclr}    &\textcolor{magenta}{19.06$\pm6.40$}     & \textcolor{blue}{43.86$\pm20.40$}     & \textcolor{magenta}{48.75$\pm6.37$}       & \textcolor{magenta}{74.95$\pm17.70$}    \\\cmidrule{2-7}
&\multirow{2}{*}{Qwen-turbo}
                     & \textsc{UBench} (Ours)               & \textcolor{magenta}{14.42$\pm7.58$}     & \textcolor{magenta}{9.58$\pm7.40$}      &  \textcolor{magenta}{54.81$\pm26.14$}   & \textcolor{magenta}{40.11$\pm22.76$}    \\
                          &&  \citet{iclr}     & \textcolor{blue}{28.70$\pm2.19$}     & \textcolor{blue}{46.25$\pm14.23$}         & \textcolor{blue}{61.31$\pm19.10$} & \textcolor{blue}{80.28$\pm5.29$}    \\\cmidrule{2-7}
&\multirow{2}{*}{Qwen-max}
                     & \textsc{UBench} (Ours)             & \textcolor{magenta}{12.51$\pm4.31$}      & \textcolor{magenta}{18.53$\pm1.03$}    & \textcolor{magenta}{36.52$\pm7.84$}    & \textcolor{magenta}{51.47$\pm8.59$}   \\
                     &&  \citet{iclr}    & \textcolor{blue}{21.09$\pm5.83$}     & \textcolor{blue}{57.39$\pm0.23$}       & \textcolor{blue}{77.50$\pm2.50$}  & \textcolor{blue}{69.17$\pm0.83$} \\ \midrule
\multirow{2}{*}{\textbf{Type}}&\multirow{2}{*}{\textbf{LLMs}} & \multirow{2}{*}{\textbf{Method}} & \multicolumn{2}{c|}{\textbf{ACE (\%)} $\downarrow$} & \multicolumn{2}{c}{\textbf{TACE (\%)} $\downarrow$} \\ \cmidrule{4-7}
                     &                        && Cosmos QA & SWAG & Cosmos QA & SWAG \\ \midrule
 \multirow{10}{*}{Open-source} 
&\multirow{5}{*}{Mistral-Instruct-7B}
                     & \textsc{UBench} (Ours)             & \textcolor{blue}{34.18$\pm3.76$}         & \textcolor{blue}{42.07$\pm2.12$}   & \textcolor{magenta}{39.49$\pm14.23$}        &  57.78$\pm23.59$    \\
                     && \citet{tengxun}         & 44.70$\pm8.96$        & 47.95$\pm9.81$    & 43.68$\pm8.53$     & \textcolor{blue}{48.65$\pm10.95$}    \\
                     && \citet{iclr}  & 43.77$\pm3.90$    &     43.25$\pm9.92$   & \textcolor{blue}{43.51$\pm9.26$ }   & \textcolor{magenta}{45.33$\pm9.98$}       \\
                      && \citet{Guo_Pleiss_Sun_Weinberger_2017}            & \textcolor{magenta}{29.78$\pm2.41$}        & \textcolor{magenta}{ 31.26$\pm5.94$}     & 50.35$\pm4.88$       & 65.54$\pm12.78$  \\
                    && \citet{gal2016dropoutbayesianapproximationrepresenting}            & 60.85$\pm6.18$        & 46.37$\pm9.27$     & 66.77$\pm2.10$       & 50.49$\pm6.37$   
                      \\\cmidrule{2-7}           
&\multirow{5}{*}{Llama3-instruct-8B}
                     & \textsc{UBench} (Ours)             & \textcolor{magenta}{14.91$\pm1.14$}        & \textcolor{magenta}{25.37$\pm5.14$}    & \textcolor{magenta}{12.36$\pm5.23$}  & \textcolor{magenta}{27.71$\pm5.51$}    \\
                     && \citet{tengxun}         & 45.26$\pm8.01$      & 42.09$\pm6.80$       & 43.25$\pm9.23$       & \textcolor{blue}{41.67$\pm8.93$}   \\
                     && \citet{iclr} & \textcolor{blue}{30.01$\pm6.80$}         & 38.42$\pm5.50$     & \textcolor{blue}{24.75$\pm5.33$}       & 45.38$\pm9.80$    \\ 
                    && \citet{Guo_Pleiss_Sun_Weinberger_2017}     & 32.99$\pm1.21$     & 47.12$\pm12.65$       & 37.64$\pm0.60$       & 54.36$\pm11.18$    \\
                     && \citet{gal2016dropoutbayesianapproximationrepresenting}     & 35.27$\pm2.79$     & \textcolor{blue}{37.85$\pm5.31$}       & 33.24$\pm5.49$       & 42.21$\pm1.34$    \\
                     \midrule
\multirow{7}{*}{Closed-source} 
&\multirow{2}{*}{GPT-3.5}
                     & \textsc{UBench} (Ours)             & \textcolor{blue}{27.06$\pm3.63$}       & \textcolor{magenta}{38.47$\pm5.00$}   & \textcolor{blue}{27.17$\pm2.01$}    & \textcolor{magenta}{22.20$\pm3.27$}    \\
                     & & \citet{iclr}     & \textcolor{magenta}{21.57$\pm3.19$}     & \textcolor{blue}{42.18$\pm13.69$}     & \textcolor{magenta}{20.95$\pm5.72$}       & \textcolor{blue}{48.43$\pm17.61$}    
                     \\\cmidrule{2-7}
&\multirow{2}{*}{Qwen-turbo}
                     & \textsc{UBench} (Ours)             &  \textcolor{magenta}{22.91$\pm1.98$}    & \textcolor{magenta}{15.47$\pm3.66$}   & \textcolor{magenta}{26.43$\pm7.38$}  & \textcolor{magenta}{13.97$\pm2.74$}    \\
                     && \citet{iclr}    & \textcolor{blue}{33.94$\pm6.56$}     & \textcolor{blue}{47.54$\pm5.87$}       & \textcolor{blue}{27.33$\pm6.03$}  & \textcolor{blue}{45.72$\pm3.73$}   \\\cmidrule{2-7}
&\multirow{2}{*}{Qwen-max}
                     & \textsc{UBench} (Ours)             &  \textcolor{magenta}{15.97$\pm2.11$}     & \textcolor{magenta}{28.42$\pm4.18$}   & \textcolor{magenta}{17.12$\pm3.32$}  & \textcolor{magenta}{22.71$\pm0.47$}  \\
                     & & \citet{iclr}    & \textcolor{blue}{31.90$\pm0.68$}     & \textcolor{blue}{54.27$\pm1.22$}      & \textcolor{blue}{27.00$\pm2.94$}  & \textcolor{blue}{61.35$\pm0.81$}   \\\midrule
\end{tabular}
\caption{We randomly sample three times, each with 100 data points, for comparison with different LLM uncertainty estimation methods. \textcolor{magenta}{Pink} represents the best, and \textcolor{blue}{blue} represents the second best. The same applies below.}
\label{mul_performance_compar}
\end{table*}

\vspace{3pt}
\noindent
\textbf{Experimental Process.} \; 
To improve reproducibility and fairness, we set the temperature of all LLMs to 0.001, except in cases where the model or API restricts this (e.g., GLM4-flash), in which case it is set to 0.01. All other parameters remain at their default values.

For all responses of LLMs, we map the chosen options to their respective confidence values. The confidence value for option $\mathcal{X}$ can be calculated as follows:
\begin{equation}  
   \mathcal{C}onf(\mathcal{X}) = (\mathcal{O}(\mathcal{\mathcal{X}}) - \mathcal{O}(\text{A})) \times 0.1 + 0.05, \label{fun:conf}
\end{equation}
where $\mathcal{C}onf(\mathcal{X})$ is the confidence value for option $\mathcal{X}$ and $\mathcal{O}(\cdot)$ represents the function that converts letter options into ASCII codes. \emph{``A''} indicates the first option from ten choices. The value of $\mathcal{X}$ ranges from A to J. As such, the value of $\mathcal{C}onf(\mathcal{X})$ is approximated to the midrange of each confidence interval, such as 5\% for interval 0\%-10\%. Finally, $\mathcal{C}onf(\mathcal{X})$ is ranging from 0.05 to    0.95. 

\subsection{Compared to Other Methods}\label{Results of UBench compared to other methods}
                            
           The comparison of \textsc{UBench}, which uses a confidence interval-based uncertainty estimation method, with other uncertainty estimation methods is shown in Table \ref{mul_performance_compar}. More experimental details are provided in Appendix \S\ref{Baseline}. We obtain the performance of \textsc{UBench} and four baseline methods on the Cosmos QA and SWAG datasets using five open-source and closed-source LLMs, by randomly sampling three times with 100 data points each. Experimental results show our method's superior performance, with a t-test indicating 77.27\% of metrics' mean values differ significantly (p < 0.05) from others. However, we also notice that \textsc{UBench} performs poorly when testing GPT-3.5 on the Cosmos QA dataset. The model's original responses indicate that this may be due to GPT-3.5's inability to properly understand the incorrect samples, resulting in overconfidence. Despite this, the optimal performance in most settings still proves the effectiveness of our method, while simultaneously reducing computational costs.

\subsection{Evaluation on Various LLMs}
With the proposed benchmark \textsc{UBench}, we primarily present the following research questions for deep analysis and discussion: 
\begin{itemize}
    \item \textbf{RQ1:} How do LLMs perform on \textsc{UBench}?
    \item \textbf{RQ2:} Do widely used prompt techniques, such as Chain-of-Thought (COT) prompt and role-playing (RP) prompt, impact the reliability of LLMs?
  
    \item \textbf{RQ3:} Does the temperature parameter affect the reliability of LLMs?
\end{itemize}
For the last two questions, we select the closed-source GPT-4 and GLM4-flash, along with the open-source GLM4-chat-9B, as the models for our research.
\subsubsection{Overall Performance}
\begin{table}[t]
\tiny
\centering
\begin{tabular}{l|c|c|c|c|>{\columncolor{gray!25}}c}

\midrule
\multirow{1}{*}{\textbf{LLMs}}  & \textbf{ECE}       & \textbf{MCE}       & \textbf{ACE}  & \textbf{TACE}   &\multirow{1}{*}{\textbf{AVG}}
\\ \midrule

    \rowcolor[HTML]{dfdfdf}
    \texttt{GPT-4o-2024-08-06} & \textcolor{blue} {13.64} & \textcolor{blue}{47.75} & \textcolor{blue}{20.90} & 21.39 & 25.92 \\
    \rowcolor[HTML]{dfdfdf}
    \texttt{Qwen-max} & 18.46 & \textcolor{magenta} {45.06} & 21.33 & 21.77 & 26.66\\
    \texttt{Yi-1.5-34B-Chat-16K} & 19.75 & 47.76 & \textcolor{magenta}{20.43} & \textcolor{blue}{20.67} & 27.15\\
    \texttt{DeepSeek V2.5} & 20.64 & 50.46 & 21.15 & \textcolor{magenta}{20.40} & 28.16 \\
    \rowcolor[HTML]{dfdfdf}
    \texttt{GLM4-flash} & 17.03 & 50.79 & 22.44 & 24.74 & 28.75 \\
    \texttt{Qwen2.5-72B-Instruct} & \textcolor{magenta}{13.00} & 54.65 & 23.18 & 25.24 & 29.02\\
    \texttt{Llama-3-8B-Instruct} & 19.65 & 52.87 & 23.16 & 21.54 & 29.31 \\
    \texttt{GLM4-chat-9B} & 17.79 & 53.78 & 23.92 & 24.86 & 30.09 \\
    \rowcolor[HTML]{dfdfdf}
    \texttt{GPT-4-1106-preview$^\ast$} & 
    17.65 & 50.43 & 24.25 & 28.21 & 30.13 \\
    \texttt{Llama-3.1-8B-Instruct} & 21.19 & 56.45 & 24.06 & 26.67 & 32.09 \\
    \rowcolor[HTML]{dfdfdf}
    \texttt{ErnieBot-v4.0$^\ast$} & 23.22 & 52.89 & 26.06 & 27.32 & 32.37 \\
    \texttt{InternLM2.5-7B-chat} 
    & 26.10 & 55.36 & 26.14 & 25.74 & 33.34 \\
    \texttt{Qwen2.5-14B-Instruct} & 27.56 & 57.31 & 26.73 & 26.31 & 34.48\\
    \texttt{InternLM2.5-20B-chat} & 22.21 & 59.03 & 26.05 & 31.07 & 34.59 \\
    \texttt{Llama-3.1-70B-Instruct} & 18.56 & 60.49 & 27.04 & 32.28 & 34.59 \\
    \texttt{Mistral-7B-Instruct-v0.2} & 22.01 & 62.89 & 27.07 & 28.14 & 35.03 \\
    \texttt{Qwen2.5-32B-Instruct} & 27.45 & 58.71 & 27.02 & 27.15 & 35.08\\
    \texttt{Qwen2.5-7B-Instruct} & 27.45 & 59.81 & 27.19 & 28.14 & 35.65\\
    \texttt{Qwen2-7B-Instruct} & 36.82 & 59.96 & 29.48 & 28.39 & 38.66\\
    \texttt{ChatGLM3-6B} & 25.23 & 70.25 & 30.13 & 37.05 & 40.67\\

    \midrule
    \textit{\textbf{TOP3 Close AVG}} & 16.38 & 47.87 & 21.56 & 22.63 & 27.11 \\
     \textit{\textbf{TOP3 Open AVG}} & 17.80 & 50.96 & 21.59 & 22.10 & 28.11 \\
     \textit{\textbf{Close AVG}} & 18.00 & 49.38& 23.00 & 24.69 & 28.77 \\
     \textit{\textbf{Open AVG}} & 23.03 & 57.32 & 25.52 & 26.91 & 33.19 \\
     \textit{\textbf{AVG}} & 21.77 & 55.34 & 24.89 & 26.35 & 32.09 \\

    \bottomrule
\end{tabular}
 \caption{LLMs' overall performance on \textsc{UBench} is evaluated using 4 metrics, with lower values indicating better performance. The final average score is calculated by combining these metrics and ranked in descending order of performance. \textit{\textbf{TOP3 Close AVG}} means the average performance of the top three closed-source models, \textit{\textbf{Close AVG}} means that of all such models, and the same goes for others. Due to resource constraints, $^\ast$ indicates that we conducted experiments on only a randomly selected one-third of the data. Please note, \textcolor{gray}{the LLM marked in gray are closed-source models.}
} 
\label{overall_performance}
\end{table}

The overall results are shown in Table \ref{overall_performance}. For additional results, please refer to Appendix \S\ref{Further Experimental Results}. Here, we report only the main findings.

\begin{figure*}[!t]
  \centering
  \includegraphics[height=0.37\linewidth, width=\linewidth]{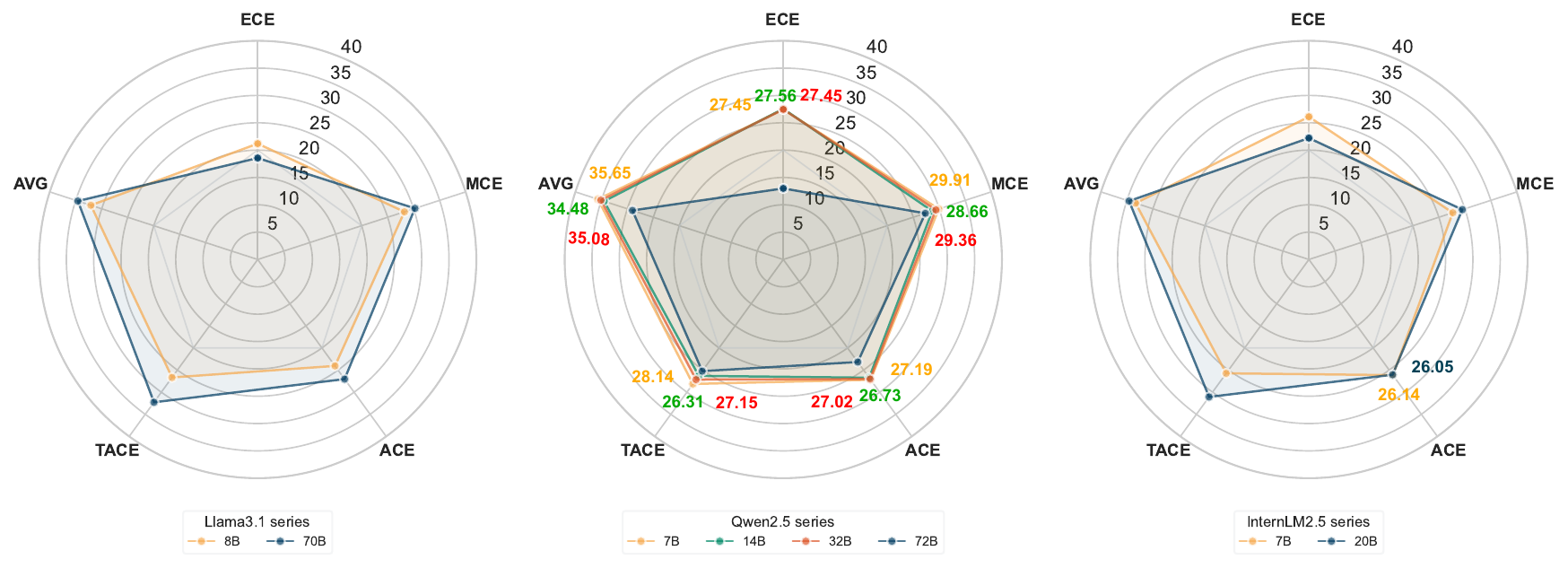}
  \caption{
    Variations in diverse metrics for the Llama3.1 series, Qwen2.5 series, and InternLM2.5 series across different model sizes. As the model size increases, both the Llama3.1 series and the InternLM2.5 series demonstrate lower reliability, which means higher uncertainty. Nevertheless, the trend for the Qwen2.5 series is not distinct. On the contrary, it attains the highest reliability when the size reaches its maximum. Please note that the MCE values in the figure have been halved for better visualization.
  }
  \label{fig:all_radar}
\end{figure*} 

\noindent
\textbf{Open Source vs. Closed Source.}  \; Among all LLMs, GPT-4o stands out for its superior performance, followed closely by Qwen-max in second place. In contrast to the top two, which are both closed-source models, the third and fourth places are occupied by two open-source models, Yi-1.5-34B and DeepSeek V2.5. In addition, the average performance difference between the TOP3 closed-source models and open-source models is only 1\%. This shows that in terms of model reliability, open-source models perform comparably to closed-source models. In the subsequent rankings, the differences between models are relatively minor, with just a 1.38\% gap between the 5th-ranked GLM4-flash and the 9th-ranked GPT-4. This range encompasses various model series, reflecting the general capabilities of LLMs developed by different companies. The next tier includes lower-performing models such as ErnieBot, InternLM, and some models from the Llama, Qwen, and GLM series. While these models may have slightly lower reliability compared to the higher-ranked models, individual models may have other factors contributing to their performance. For example, Llama-3.1-70B achieved an impressive score of 18.56 on the ECE metric, significantly outperforming models in the same tier. However, its MCE score is relatively high, suggesting that its responses may frequently display extreme levels of confidence or uncertainty.

\noindent
\textbf{Across different model size.} \; As shown in Figure~\ref{fig:all_radar}, it is worth noting that there is no clear correlation between the model size and the reliability of the model. For instance, for the Llama3.1 and InternLM2.5 series models, the larger the model, the greater the uncertainty and the lower the reliability. However, this rule does not apply to the Qwen2.5 series. Even the reliability of the Qwen2.5-72B model is much higher than that of the Qwen2.5-7B model. This seems to imply that the model size is not the only factor affecting the model reliability.
Furthermore, when comparing models across different series, such as Llama-3-8B and GLM4-9B, we find that they exhibit reliability that surpasses many models with parameter counts exceeding 10B. A possible explanation for this discrepancy is that some LLMs prioritize improving performance across various tasks, while others not only enhance task performance but also address areas such as hallucination, safety, and other capabilities that contribute to greater reliability. These findings highlight the importance of incorporating uncertainty estimation into LLM research and evaluation systems.

\noindent
\textbf{Across different metrics.} \; In particular, Qwen2.5-72B is the top-performing model for ECE, closely followed by GPT-4o. While the difference between Qwen2.5-72B and GPT-4o is marginal, both models significantly outperform Qwen-max and Yi-1.5, which are ranked 2nd and 3rd, by 5.46\% and 6.11\%. However, Qwen2.5-72B's average ranking is not exceptional, similar to Llama-3.1-70B. Both models struggle with MCE, ACE, and TACE. On one hand, this suggests that the two models may be overly sensitive to extreme values. On the other hand, this result reflects the strong complementarity of MCE, ACE, and TACE to ECE in extreme and multi-classification scenarios, demonstrating the scientificity of \textsc{UBENCH}'s use of average metrics for model ranking. The ranking of ACE scores is largely consistent with the ranking of average values. Among them, Yi-1.5 stands out most prominently, followed by GPT-4o. When the confidence threshold is set to 0.5, the open-source DeepSeek performs best in this metric. Moreover, the top four models in terms of average performance do not have a significant gap among them. This indicates that the reliability of these models can remain relatively high even in scenarios of high confidence.

\begin{table}[htbp]
\centering
\scriptsize
\begin{tabular}{
   l|c|c|c|c|>{\columncolor{gray!25}}c}
\toprule 
\multirow{1}{*}{\textbf{LLMs}}& \textbf{ECE}       & \textbf{MCE}       & \textbf{ACE}  & \textbf{TACE}   &\multirow{1}{*}{\textbf{AVG}}    
                               \\   \midrule

    \texttt{GPT-4}& 
    17.65 & 50.43 & 24.25 & 28.21 & 30.13 \\
    \texttt{w/o CoT}  & 20.55 & 47.30 & 23.70 & 29.63 & 30.29  \\

    \texttt{w/o RP}  & 17.06 & 56.07 & 26.20 & 31.61 & 32.74  \\
    
    \texttt{w/o CoT\&RP}& 21.15 & 50.14 & 24.96 & 32.08 & 32.08 \\

    \midrule
    
    \texttt{GLM4-flash} & 17.03 & 50.79 & 22.44 & 24.74 & 28.75 \\
    
    \texttt{w/o CoT} & 26.65 & 53.25 & 25.90 & 29.74 & 33.89   \\
    
    \texttt{w/o RP} & 19.45 & 52.02 & 23.44 & 24.31 & 29.80    \\

    \texttt{w/o CoT\&RP}& 29.87 & 63.04 & 32.19 & 33.52 & 39.65 \\
    
    \midrule
    
    \texttt{GLM4} & 17.79 & 53.78 & 23.92 & 24.86 & 30.09\\
    
    \texttt{w/o CoT} & 26.76 & 54.13 & 26.15 & 28.85 & 33.97 \\
    
    \texttt{w/o RP} & 19.62 & 51.40 & 23.37 & 24.07 & 29.61   \\

    \texttt{w/o CoT\&RP}& 28.71 & 55.49 & 27.80 & 27.60 & 34.90 \\

      \bottomrule
\end{tabular}
 \caption{LLMs' performance on \textsc{UBench} with different prompt changes. \textbf{"w/o"} means removing the prompt, \textbf{"CoT"} refers to the Chain-of-Thought prompt, and \textbf{"RP"} represents the role-playing prompt (the same applies below).} 
\label{no_cot}
\end{table}

\subsubsection{Effects of CoT and RP Prompts}
The exploratory experiment results with CoT and RP prompt as variables are presented in Table \ref{no_cot} and in Figure \ref{fig:cot}. 
Overall, the use of CoT and RP has a positive impact on the reliability of all three models. The only exception is that when RP is removed alone, the average score of GLM4 decreases slightly by 0.48\%. Otherwise, whether removing CoT, RP, or both, the average scores of the three models remain stable or increase, indicating a decrease in reliability. Among them, removing CoT alone results in an average increase of 3.06\% in the scores of the three models, which is higher than the 1.06\% increase when removing RP alone, indicating that CoT has a greater impact than RP. Removing both CoT and RP results in an average increase of 5.62\% in the model scores, demonstrating the effectiveness and good synergy of both in improving model reliability.

Looking at the models individually, GPT-4 appears to be less sensitive to CoT. Specifically, results on GPT-4 show that removing CoT alone has a similar performance to not removing it, and when both CoT and RP are removed, the model's score decreases compared to removing only RP, indicating an improvement in reliability. This may be because GPT-4 itself has strong reasoning and problem-solving abilities. Moreover,  GPT-4's training data may already contain many reasoning and thinking processes similar to CoT, allowing it to internally simulate thought chains without explicit CoT instructions. In contrast, GLM4 and GLM4-flash can benefit more from CoT. However, RP does not seem to have a significant impact on GLM4 and GLM4-flash, possibly because role-playing prompts require a high level of ability to maintain specific role characteristics and language styles. GLM4 and GLM4-flash's capabilities may not meet this requirement.

\subsubsection{Effects of Temperature}
\begin{table}[t]
\scriptsize
\centering
\begin{tabular}{
   l|c|c|c|c|>{\columncolor{gray!25}}c}
\toprule 
\multirow{1}{*}{\textbf{LLMs}}  & \textbf{ECE}       & \textbf{MCE}       & \textbf{ACE}  & \textbf{TACE}   &\multirow{1}{*}{\textbf{AVG}}
                              
                               \\ \midrule

    \texttt{GPT-4}  & 17.25 & 44.94 & 22.13 & 22.87 & 26.80 \\

    \texttt{w/ 0.4}  & 18.74 & 53.04 & 24.04 & 27.99 & 30.95 \\

    \texttt{w/ 0.8}  & 18.57 & 54.98 & 25.18 & 28.94 & 31.92 \\
    
    \texttt{w/ 1.2} & 18.99 & 54.50 & 25.54 & 30.90 & 32.48 \\

    \texttt{w/ 1.6} & 20.28 & 64.73 & 29.39 & 31.55 & 36.49 \\

    \texttt{w/ 2.0} & 21.63 & 64.58 & 29.35 & 29.40 & 36.24 \\

    \midrule

    \texttt{GLM4-flash} & 17.03 & 50.79 & 22.44 & 24.74 & 28.75 \\

    \texttt{w/ 0.2} & 17.56 & 46.64 & 21.50 & 22.70 & 27.10 \\

    \texttt{w/ 0.4} & 17.83 & 40.84 & 19.76 & 21.68 & 25.03 \\
    
    \texttt{w/ 0.6}  & 17.98 & 38.37 & 19.22 & 20.84 & 24.10\\

    \texttt{w/ 0.8} & 16.74 & 35.58 & 16.82 & 19.95 & 22.27  \\

    \texttt{w/ 1.0} & 17.04 & 37.16 & 17.14 & 19.52 & 22.72 \\
    \midrule

    \texttt{GLM4} & 17.79 & 53.78 & 23.92 & 24.86 & 30.09  \\

    \texttt{w/ 0.4} & 18.39 & 43.00 & 20.03 & 22.13 & 25.89 \\

    \texttt{w/ 0.8} & 17.39 & 36.76 & 16.81 & 20.11 & 22.77 \\
    
    \texttt{w/ 1.2}  & 16.81 & 35.52 & 16.88 & 19.58 & 22.20\\

    \texttt{w/ 1.6} & 16.65 & 31.40 & 16.50 & 18.29 & 20.71 \\

    \texttt{w/ 2.0} & 16.55 & 32.76 & 16.63 & 18.26 & 21.05 \\

    \bottomrule
\end{tabular}
 \caption{LLMs' performance at different temperature settings on \textsc{UBench}. Due to computational resource limitations, the results of GPT-4 in this table are based on a randomly selected one-fifth of the sample data.} 
\label{table_temperate}
\end{table}

We study the changes in the reliability of LLMs within the temperature range of 0 to 2, using an interval of 0.4. Due to the temperature range limitations of the GLM4-flash API, which only allows values between 0 and 1, GLM4-flash is studied using intervals of 0.2. The performance of GPT-4, GLM4-flash, and GLM4 at different temperatures is shown in Table \ref{table_temperate} and in Figure \ref{fig:tem}. Experimental results show that GPT-4's reliability decreases with rising temperatures, while GLM4-flash and GLM4's reliability increases. The trends of the four evaluation metrics are consistent. From the original responses of the models, it can be seen that as the temperature rises, the replies of LLMs become more random, which may not be conducive to their selection of accurate confidence options. However, with an increase in temperature, the responses of GLM4-flash and GLM4 incorporate more reasoning processes (please refer to Table~\ref{temperature_glm4_case} and Table~\ref{temperature_glm4_flash_case} for specific case). This helps them select more appropriate confidence options and reduces the uncertainty of their responses. In contrast, GPT-4 has stronger reasoning and problem-solving abilities than GLM4-flash and GLM4. Therefore, this phenomenon is not as obvious in GPT-4, so its reliability decreases with the increase in temperature. It should be noted that high temperature may lead to LLMs' outputs becoming chaotic, thereby restricting their applicability and increasing the randomness of the experiment.

\begin{table}[]
\scriptsize
\centering
\begin{tabular}{
   l|c|c|c|c|>{\columncolor{gray!25}}c}
\toprule 
\multirow{1}{*}{\textbf{Setting}}  & \textbf{ECE}       & \textbf{MCE}       & \textbf{ACE}  & \textbf{TACE}   &\multirow{1}{*}{\textbf{AVG}}
                              
                               \\ \midrule

\texttt{5 intervals} & 29.72 & 34.33 & 17.82 & 1.95 & 20.95 \\
\texttt{10 intervals} & 23.22 & 34.05 & 9.51 & 5.38 & 18.04 \\
\texttt{20 intervals} & 31.45 & 39.80 & 16.56 & 14.46 & 25.57 \\

    \bottomrule
\end{tabular}
     \caption{Performance comparison of Qwen2.5-7B-Instruct on the SWAG dataset using different numbers of intervals (5, 10, and 20).}
    \label{num_intervals}
\end{table}

\subsection{Ablation Studies on Confidence Intervals}
To further investigate our method, we conduct an ablation study on the number of intervals. As shown in Table~\ref{num_intervals}, We conduct experiments on the SWAG dataset using Qwen2.5-7B-Instruct with different numbers of intervals (5, 10, and 20). The results show that using 10 intervals achieves the best performance, further validating the rationality of our interval choice. We believe that using more or fewer intervals would affect the model's expressive capacity, which is not conducive to capturing uncertainty effectively.

\section{Case Study} \label{Case Study}
Since our methods are heuristic in nature, we attempt to illustrate the advantages of our approach relative to the method by \citet{iclr} (hereafter referred to as the hybrid approach) by example. Tables \ref{tab-xiong} and 
\ref{tab-our} show the responses of Mistral-7B to a question from the Cosmos QA dataset based on these two methods. Due to the length of the text, we only present three answers generated by the hybrid approach. Firstly, our method is characterized by shorter questions and answers, reflecting its conciseness and low resource consumption. Content-wise, with our prompt strategy (refer to Table \ref{tab-our}), the model assigns a confidence level of 80\%-90\% to the correct answers and only 0-10\% to the incorrect ones. This indicates that Mistral-7B has a good understanding of the prompt and fully demonstrates its comprehension and reasoning abilities. In contrast, among the three answers obtained using the hybrid approach (refer to Table \ref{tab-xiong}), one incorrectly selected option B (None of the above choices) with a 70\% confidence level, indicating that Mistral-7B's understanding of the question was incomplete. In the other two correct answers, the model only provided a 60\%-70\% confidence level. The model exhibits a higher degree of uncertainty for the same question. Its performance is also further away from human expectations, which is detrimental to its application.

\section{Conclusion}
Focusing on the assessment of reliability in LLMs, we present \textsc{UBench}, a new benchmark for uncertainty estimation in LLMs based on multiple choice questions. The benchmark consists of 12K ten-choice questions in four categories: knowledge, language, understanding, and reasoning. Comparative experimental results show that our confidence interval-based method outperforms other SOTA uncertainty estimation methods. Additionally, We assess the reliability of 20 mainstream LLMs, which include both open and closed sources, on this benchmark. We reveal that even the most advanced LLMs still exhibit low reliability in their predictions, especially in extreme cases, which pose potential risks. Therefore, it is necessary to incorporate uncertainty estimation into the evaluation of LLMs. Further exploratory analysis reveals that the incorporation of CoT and role-playing prompt methods generally benefits LLMs in demonstrating stronger reliability, whereas changes in temperature have varying effects on different LLMs. Additionally, we conduct a case study to demonstrate the effectiveness of our proposed method. We hope that this study will play an important role in the further development and application of LLMs.

\section*{Limitations}
Our work is a new attempt to measure the uncertainty of LLMs by constructing benchmarks containing ten multiple-choice questions and to explore potential factors that may affect their reliability. Although our work provides a comprehensive uncertainty assessment of LLMs and compares it with other uncertainty estimation methods while analyzing potential effect factors, some limitations remain. These limitations may guide our future work.

First, we assess the reliability of LLMs in the four main abilities of knowledge, language, understanding, and reasoning. However, the abilities of LLMs encompass more than these, and the development of more extensive tests designed to assess the reliability of LLMs is necessary. One direction to focus on is to evaluate the reliability of LLMs in multimodal scenarios \cite{yin2023survey}.

Second, similar to other well-known works, our work is based on multiple-choice questions~\cite{zhang2023safetybench, chen2024tombench}. While these questions offer advantages like standardization and ease of evaluation, they require the model to follow instructions and may be affected by positional bias in the answer options~\cite{zheng2023large}. This is an important area for future exploration.

Last but not least, we explore the effects of CoT prompt, role-playing prompt and temperature on the reliability of LLMs. However, there are many other potential factors affecting the reliability of LLMs, such as model fine-tuning, model quantification, etc., which deserve further exploration.

\section*{Ethics Statement}
This work is based on experiments conducted with several models and datasets, which are widely used for scientific research and do not pose potential disputes. For closed-source models, we have paid for access to their APIs, and the access frequency remains within normal limits. Additionally, \textsc{UBench} does not include prompts that could trigger harmful outputs from LLMs, making it difficult for potential attackers to exploit the questions in \textsc{UBench} to induce detrimental responses. During the data validation process, we have maintained clear and proactive communication with volunteers, ensuring their voluntary participation and eliminating any potential risks.

\section*{Acknowledgements}
We sincerely thank all the anonymous reviewers
for providing valuable feedback. This work is supported by the National Natural Science Foundation
of China (No.62406151).

\bibliography{anthology,custom}
\appendix

\begin{table*}[htbp]
\centering
\small
\setlength{\fboxrule}{1pt}
\setlength{\fboxsep}{10pt} 
\adjustbox{minipage=\linewidth}{
    \fbox{
        \begin{tabular}{p{\linewidth}}
       
\textbf{[Prompt]:} \\
Read the multiple-choice question, analyze step by step, select the correct option and give option letter e.g. A or B as your answer. Use the following format to provide your answer and confidence level: Explanation: [insert step-by-step analysis here] Answer and Confidence (0-100): [Your answer, please only include the capital letter, e.g. B], [Your confidence level, please only include the numerical number, e.g. 80]\% \\
Note: The confidence level indicates the degree of certainty you have about your answer and is represented as a percentage. For instance, if your confidence level is 80\%, it means you are 80\% certain that your answer is correct and there is a 20\% chance that it may be incorrect. \\
\textbf{[Question]:} \\
What may happen if Alexis Cruz 's plan succeeded then someone pushed it even further ? \\
Context: That was nice . Garret Wang showed up and Alexis Cruz was with us . He tried to stick a cork in his nose , but it would n't fit \\
A. a nasal fracture may happen . \textcolor{red}{(True)} \\
B. None of the above choices . \\
C. cork will go to his brain eventually . \\
D. cork will explode to bits in his nose . \\

\textbf{[Response 1]:} \\
To analyze this question, we need to consider the context and the possible outcomes of the scenario described.
Understanding the Scenario: Alexis Cruz's plan involves some action, but the specifics are not provided. We need to infer the potential consequences based on the given options.
Evaluating the Options:
A. a nasal fracture may happen: This suggests a physical injury, specifically a broken nose.
B. None of the above choices: This option implies that none of the other listed outcomes are likely.
C. cork will go to his brain eventually: This is a highly unlikely and extreme outcome, as it would require the cork to penetrate the nasal cavity and enter the brain, which is not a typical or realistic consequence of most actions.
D. cork will explode to bits in his nose: This is also an unlikely outcome, as corks do not typically explode under normal circumstances.
Logical Deduction: Without specific details about Alexis Cruz's plan, we can only consider the most plausible outcomes. Option A (a nasal fracture) is a realistic possibility if the plan involves physical force or impact to the nose. Options C and D are highly implausible. Option B suggests that none of the other options are correct, but without more context, it's difficult to definitively rule out Option A.
Given the lack of specific details, the most plausible answer is:
Explanation: Option A is a realistic possibility involving physical injury, while Options C and D are highly unlikely. Option B could be correct if the plan does not involve any risk of nasal fracture, but without more context, Option A is the most plausible.
Answer and Confidence: A, 60\% \\

\textbf{[Response 2]:} \\
To analyze this question, we need to consider the context and the possible consequences of the actions described. \\
Alexis Cruz's plan: Without specific details about Alexis Cruz's plan, we can't determine the exact outcome. However, we can infer that the plan involves some risky or dangerous activity, given the options provided.
Pushing the plan further: This implies escalating the risk or intensity of the original plan. \\
Evaluating the options: \\
A. a nasal fracture may happen: This suggests a physical injury, specifically to the nose. \\
B. None of the above choices: This option denies all the other possibilities. \\
C. cork will go to his brain eventually: This is a highly unlikely and extreme scenario. \\
D. cork will explode to bits in his nose: This is also an extreme and unlikely scenario. \\ 
Given the lack of specific details about Alexis Cruz's plan, it's difficult to definitively choose any of the extreme or unlikely options (C or D). Option A is a possible outcome if the plan involves some form of physical risk to the nose, but without more context, we can't be certain. Therefore, the safest and most logical choice is: \\ Answer and Confidence (0-100): B, 70\%

\textbf{[Response 3]:} \\
To analyze this question, we need to consider the context and the possible outcomes of the scenario described: "What may happen if Alexis Cruz's plan succeeded then someone pushed it even further?" Understanding the Scenario: Without specific details about Alexis Cruz's plan, we can infer that it involves some risky or dangerous activity, given the outcomes mentioned in the options. \\
Evaluating the Options: \\
A. a nasal fracture may happen: This suggests a physical injury, specifically a broken nose. \\
B. None of the above choices: This option implies that none of the other listed outcomes would occur. \\
C. cork will go to his brain eventually: This suggests a severe and potentially fatal outcome, where an object (cork) enters the brain. \\
D. cork will explode to bits in his nose: This suggests a violent and dangerous outcome where an object (cork) explodes inside the nose. \\
Logical Deduction: \\
Option A (nasal fracture) is a plausible outcome of a risky activity involving the nose. \\
Option C (cork going to the brain) is highly unlikely and would require extreme and specific circumstances. \\
Option D (cork exploding in the nose) is also highly unlikely and would require specific conditions that are not typically associated with common activities. \\
Option B (none of the above) could be correct if the plan and its escalation do not lead to any of the specific outcomes listed. \\ 
Given the lack of specific details about Alexis Cruz's plan, the most plausible outcome among the options provided is a nasal fracture, as it is a common injury that could result from various risky activities.\\
Explanation: A nasal fracture is a plausible outcome of a risky activity involving the nose, making it the most likely choice among the options provided.\\ 
Answer and Confidence: A, 70\%

        \end{tabular}
    }
}
\caption{~\citet{iclr} uses Mistral-Instruct-7B as the subject in a case study on the Cosmos QA dataset.}
\label{tab-xiong}
\end{table*}
\begin{table*}[htbp]
\centering
\small
\setlength{\fboxrule}{1pt} 
\setlength{\fboxsep}{10pt} 
\adjustbox{minipage=\linewidth}{
    \fbox{
        \begin{tabular}{p{\linewidth}}
        
\textbf{[Question]:} \\
What may happen if Alexis Cruz 's plan succeeded then someone pushed it even further ? \\
Context: That was nice . Garret Wang showed up and Alexis Cruz was with us . He tried to stick a cork in his nose , but it would n't fit .\\
\midrule \\
\textbf{[Candidate Answer]:} a nasal fracture may happen . \textcolor{red}{(True)}

\textbf{[Response]:} \\
If Alexis Cruz's plan to stick a cork in his nose had succeeded and someone pushed it even further, the likelihood of a nasal fracture would be quite high due to the delicate nature of the nasal bones and the potential force applied. Therefore, the estimated probability would be:I: 80\%-90\% \\ \\

\midrule \\

\textbf{[Candidate Answer]:} cork will go to his brain eventually . \textcolor{red}{(False)} \\

\textbf{[Response]:} \\
The scenario described is highly unlikely and extremely dangerous. The nasal cavity is not directly connected to the brain in a way that a cork could simply travel upwards into the brain. However, attempting such an action could cause serious harm, such as damaging nasal passages, causing infection, or potentially leading to other complications if the object were to become lodged or forcefully pushed.
Given the extreme nature and low likelihood of the cork actually reaching the brain, the probability would be very low. Therefore, the most appropriate option would be:
A: 0-10\% \\

        \end{tabular}
    }
}
\caption{\textsc{UBench} (ours) uses Mistral-Instruct-7B as the subject in a case study on the Cosmos QA dataset.}
\label{tab-our}
\end{table*}

\section{More Related Work} \label{More Related Work}
Apart from the main method categories mentioned in~\S\ref{Uncertainty Estimation for LLMs}, there are also similarity-based and domain-specific metrics. Similarity-based approaches rely on coherence between generations. SelfCheckGPT \cite{manakul2023selfcheckgpt} detects hallucinations via sampling and coherence checks but incurs high computational costs. Domain-specific benchmarks are designed for specific tasks, which include \citet{vedantam2021curi} for compositional reasoning under uncertainty, \citet{Kirchhof2023URLAR} for uncertainty-aware representation learning, \citet{zablotskaia2023uncertainty} for summarization benchmarks and \citet{li2024uncertainty} for code generation. Although these research achievements focus on niche applications, they still require downstream adaptation and optimization. 

\section{Prompt and Empirical Validation}

\subsection{Prompt Templates}\label{Prompt Templates}
\begin{table*}[htbp]
\centering
\small
\setlength{\fboxrule}{1pt}
\setlength{\fboxsep}{10pt}
\adjustbox{minipage=\linewidth}{
    \fbox{
        \begin{tabular}{p{0.95\linewidth}}
       
You are \textbf{\{specific type of role\}}. \\ \\

You will see \textbf{\{task-specific fields\}}, a question, an answer, and ten confidence interval options. Your task is to analyze the given \textbf{\{task-specific fields\}}, question, and answer, and select the confidence interval that you believe reflects the correctness of the answer. \\ \\

For each example, you must carefully read the given \textbf{\{task-specific fields\}}, question, and answer, then analyze the answer and systematically consider the likelihood of its correctness. Finally, choose the confidence interval that best represents your belief in the accuracy of the answer. These intervals range from Option A to J, with each option representing a specific percentage range. Please think step by step. \\ \\

Your answer must follow the format specified below: \\
Answer format: \\
Answer: [Option], e.g., Answer: [C] \\

Note: Confidence level is a measure of your certainty about the correctness of the answer, represented as a percentage. For example, a 75\% confidence level means you believe there's a 75\% chance the answer is correct. \\ \\
\textbf{\{Task-specific fields\}} \\
Question: \textbf{\{Question\}} \\
Answer: \textbf{\{Answer\}}  \\
Options: \\
A: 0-10\% \\
B: 10\%-20\% \\
C: 20\%-30\% \\
D: 30\%-40\% \\
E: 40\%-50\% \\
F: 50\%-60\% \\
G: 60\%-70\% \\
H: 70\%-80\% \\
I:  80\%-90\% \\
J:  90\%-100\% \\
        \end{tabular}
    }
}
\caption{Evaluating the prompt template for LLMs.}
\label{tab:prompt}
\end{table*}
\begin{table*}[htbp]
\centering
\small
\setlength{\fboxrule}{1pt} 
\setlength{\fboxsep}{10pt}
\adjustbox{minipage=\linewidth}{
    \fbox{
        \begin{tabular}{p{0.95\linewidth}}
       
I want you to act as an uncertainty benchmark generator. \\ \\

Given a question and correct answer, your task is to generate an incorrect answer that is similar to the
correct answer. \\ \\

Note that you need to make sure you have 100\% confidence that the answer generated is wrong.   \\
Please provide the incorrect answer directly, without any other words. \\ \\
Here is an example:

Question: Dana can run at a rate of speed four times faster than she can walk, but she can skip at a rate of speed that is half as fast as she can run. If she can skip at 3 miles per hour, how many miles can she travel in six hours if she spends one-third of the time running and two-thirds of the time walking?

Correct answer: 18

Incorrect answer: 12 \\ \\

Now generate an incorrect answer for the following question.

Question: \textbf{\{Question\}}

Correct answer: \textbf{\{Answer\}}
        \end{tabular}
    }
}
\caption{Prompt template for generating negative samples.}
\label{tab:gen_prompt}
\end{table*}
The prompt template for evaluating LLMs' reliability is shown in Table~\ref{tab:prompt}. 
The prompt template for generating negative samples is shown in Table~\ref{tab:gen_prompt}. 

\subsection{Empirical Validation on Prompt Design} \label{Empirical validation on prompt design}
\begin{table*}[htbp]
\centering
\tiny
\resizebox{\textwidth}{!}{%
\begin{tabular}{
   c|c|c|c|c|c|>{\columncolor{gray!25}}c}
\toprule 
{\textbf{Method}} & \multirow{1}{*}{\textbf{Prompt type}}  & \textbf{ECE (\%)} $\downarrow$ & \textbf{MCE (\%)} $\downarrow$ & \textbf{ACE (\%)} $\downarrow$ & \textbf{TACE (\%)} $\downarrow$ & \multirow{1}{*}{\textbf{AVG (\%)} $\downarrow$} \\ 
\midrule 

\multirow{2}{*}{Max softmax logit} 
& \texttt{Standard Prompt} & \textcolor{blue}{25.07$\pm3.26$} & \textcolor{blue}{91.61$\pm11.86$} & \textcolor{blue}{34.21$\pm5.17$} & \textcolor{blue}{53.00$\pm8.88$} & \textcolor{blue}{50.97$\pm7.29$} \\
& \texttt{Interval-based prompt (Ours)} & \textcolor{magenta}{17.76$\pm2.83$} & \textcolor{magenta}{67.01$\pm18.38$}  & \textcolor{magenta}{26.59$\pm0.99$} & \textcolor{magenta}{21.69$\pm7.27$}  & \textcolor{magenta}{33.26$\pm7.37$} \\
\midrule

\multirow{2}{*}{Entropy} 
& \texttt{Standard Prompt} & \textcolor{magenta}{18.83$\pm1.05$} & \textcolor{blue}{98.13$\pm1.32$} & \textcolor{blue}{35.67$\pm4.06$} & \textcolor{blue}{59.10$\pm12.58$} & \textcolor{blue}{52.93$\pm4.75$} \\
& \texttt{Interval-based prompt (Ours)} & \textcolor{blue}{23.88$\pm3.98$} & \textcolor{magenta}{78.56$\pm18.92$}  & \textcolor{magenta}{31.16$\pm3.97$} & \textcolor{magenta}{40.03$\pm15.38$}  & \textcolor{magenta}{43.41$\pm10.56$} \\
\midrule

\multirow{2}{*}{Margin} 
& \texttt{Standard Prompt} & \textcolor{blue}{19.24$\pm2.87$} & \textcolor{blue}{92.73$\pm10.28$} & \textcolor{blue}{35.77$\pm7.62$} & \textcolor{blue}{62.97$\pm11.68$} & \textcolor{blue}{52.68$\pm8.11$} \\
& \texttt{Interval-based prompt (Ours)} & \textcolor{magenta}{11.59$\pm1.16$} & \textcolor{magenta}{62.98$\pm24.72$}  & \textcolor{magenta}{17.62$\pm3.52$} & \textcolor{magenta}{18.28$\pm4.64$}  & \textcolor{magenta}{27.62$\pm8.51$} \\

    \bottomrule    
    \end{tabular}
    }

    \caption{Performance of various prompt types on three uncertainty quantification methods, tested with Mistral-Instruct-7B on the SWAG dataset over three sampling iterations (100 samples per iteration).} 
    \label{prompt_validation}
\end{table*}

As shown in Table~\ref{prompt_validation}, we conduct experiments using Mistral-Instruct-7B on the SWAG dataset with three sampling iterations (settings consistent with~\S\ref{Results of UBench compared to other methods}), comparing the standard prompt and interval-based prompt across three logit-based uncertainty quantification methods. For the standard prompt, we utilize the prompt from~\citet{iclr}. The results reveal that the interval-based prompt significantly outperforms the standard prompt, further substantiating the efficacy of our proposed approach.

\section{Baseline Experimental Setup}\label{Baseline}
To ensure fairness, in all the baseline experiments, the temperature parameter is set to 0.001. For the method proposed by~\citet{iclr}, we perform 5 times sampling. For the method proposed by~\citet{tengxun}, we set the error rate $\alpha$ to 0.1. These are the default optimal settings. For Temperature Scaling~\cite{Guo_Pleiss_Sun_Weinberger_2017}, to determine the optimal temperature, we sample 500 examples from the training sets of Cosmos QA and SWAG, searching within the range [0.001, 5], with up to 10 search attempts. The final temperatures selected are 4.72 for Llama and 4.96 for Mistral. For Monte-Carlo Dropout~\cite{gal2016dropoutbayesianapproximationrepresenting}, we set the dropout rate to 0.2 and perform 5 inference iterations during the ensemble process.

\section{Datasets, Models, and Metrics}
\subsection{Datasets Overview} \label{Datasets Overview}
\begin{table}[htbp]
\centering

\tiny
\begin{tabular}{lll}
\toprule
Category & Dataset & Subset\\
\midrule

\multirow{4}{*}{\includegraphics[width=0.5cm]{figs/icon/Language.png}Language} & WIC~\citep{WIC} & Dev\\
& WSC~\citep{WSC} & Dev\\
& COLA~\citep{COLA} & Dev\\
& QQP\footnote{\url{https://www.kaggle.com/c/quora-question-pairs}} & Dev \\
\midrule

\multirow{5}{*}{\includegraphics[width=0.5cm]{figs/icon/Knowledge.png}Knowledge} & CommonSenseQA~\citep{CommonsenseQA} & Dev  \\
& TruthfulQA~\citep{TruthfulQA} & Dev  \\
& popqa-parents-lying\footnote{\url{https://huggingface.co/datasets/atmallen/popqa-parents-lying}} & Test\\
& NQ~\citep{NQ} & Test \\
& WikiQA~\citep{wikiqa} & Test \\

\midrule

\multirow{8}{*}{\includegraphics[width=0.55cm]{figs/icon/Understanding.png}Understanding} & RACE~\citep{RACE} & Test \\
& MMLU~\citep{MMLU} & Test \\
& LogiQA~\citep{LogiQA} & Test \\
& SWAG~\citep{SWAG} & Dev \\
& Cosmos QA ~\citep{CosmosQA} & Dev \\
& Social IQA~\citep{SocialIQA} & Dev \\
& DialogSum~\citep{DialogSum} & Test \\
& CommonGen~\citep{CommonGen} & Dev \\
\midrule
\multirow{7}{*}{\includegraphics[width=0.5cm]{figs/icon/Reasoning.png}Reasoning} & GSM8K~\citep{GSM8K} & Test\\
& AQuA~\citep{AQuA} & Test\\
& AddSub~\citep{AddSub} & Test \\
& MultiArith~\citep{MultiArith} & Test\\
& SingleEq~\citep{SingleEQ} & Test \\
& SVAMP~\citep{SVAMP} & Test \\
& PIQA~\citep{PIQA} & Dev \\
\bottomrule
\end{tabular}
\caption{Datasets overview.}
\label{tab:datasets}
\end{table}
An overview of the open source datasets used in \textsc{UBench} is shown in  Table \ref{tab:datasets}. For datasets where the test subset is fully accessible, we use the test subset; otherwise, we use the dev subset.

\subsection{LLMs in Experiment}

The models evaluated in our experiment are shown as Table~\ref{tab:LLMs}. Please note that our experiments are conducted on the original model without fine-tuning.

\begin{table*}[htbp]
    \centering
    \footnotesize
    \renewcommand{\arraystretch}{1.0}
    \begin{tabular}{lccccc}
    \toprule
    \textbf{Model} & \textbf{Parameters} & \textbf{Access} & \textbf{Version} & 
    \textbf{Language} & \textbf{Publisher} \\
    \midrule
    \href{https://openai.com/index/gpt-4o-system-card/}{\texttt{GPT-4o} \citep{OpenAI_2023}} & undisclosed & API & 2024-08-06 & zh/en & \multirow{2}{*}{OpenAI}    \\
    \href{https://openai.com/gpt-4}{\texttt{GPT-4} \citep{OpenAI_2023}} & undisclosed & API & 1106-preview & zh/en   \\

    \midrule
    \href{https://cloud.baidu.com/wenxin.html}{\texttt{ErnieBot} \citep{Ernie}} & undisclosed & API & v4.0 & zh/en & Baidu \\
    \midrule
    \href{https://www.deepseek.com/}{\texttt{DeepSeek} \citep{deepseekv2}} & 236B & API & v2.5 & zh/en & DeepSeek \\
    \midrule
    \href{https://help.aliyun.com/zh/dashscope/api-reference}{\texttt{Qwen-max} \citep{QWEN}} & undisclosed & API & 1201 & zh/en& \multirow{6}{*}{Alibaba Cloud}   \\
    \href{https://huggingface.co/Qwen/Qwen2-7B-Instruct}{\texttt{Qwen2-Instruct} \citep{qwen2}} & 7B & Weights & v2.0 & zh/en   \\
     \href{https://huggingface.co/Qwen/Qwen2.5-7B-Instruct}{\texttt{Qwen2.5-Instruct} \citep{qwen2.5}} & 7B & Weights & v2.5 & zh/en   \\
        \href{https://huggingface.co/Qwen/Qwen2.5-14B-Instruct}{\texttt{Qwen2.5-Instruct} \citep{qwen2.5}} & 14B & Weights & v2.5 & zh/en   \\
     \href{https://huggingface.co/Qwen/Qwen2.5-32B-Instruct}{\texttt{Qwen2.5-Instruct} \citep{qwen2.5}} & 32B & Weights & v2.5 & zh/en   \\
     \href{https://huggingface.co/Qwen/Qwen2.5-72B-Instruct}{\texttt{Qwen2.5-Instruct} \citep{qwen2.5}} & 72B & Weights & v2.5 & zh/en   \\
    \midrule
  \href{https://ai-bot.cn/bigmodel/}{\texttt{GLM4-flash} \citep{glm2024chatglm}} &undisclosed & API & - & zh/en & \multirow{3}{*}{Tsinghua \& Zhipu} \\
    \href{https://huggingface.co/THUDM/chatglm3-6b}{\texttt{ChatGLM3} \citep{GLM}} & 6B & Weights & v3.0 & zh/en \\
    \href{https://huggingface.co/THUDM/glm-4-9b-chat}{\texttt{GLM4-chat} \citep{glm2024chatglm}} & 9B & Weights & v4.0 & zh/en & \\
    \midrule

    \href{https://huggingface.co/meta-llama/Meta-Llama-3-8B-Instruct}{\texttt{Llama3-Instruct} \citep{llama3modelcard}} & 8B & Weights & v3.0 & en & \multirow{3}{*}{Meta AI}  \\
    \href{https://huggingface.co/meta-llama/Meta-Llama-3.1-8B-Instruct}{\texttt{Llama3.1-Instruct} \citep{llama3modelcard}} & 8B & Weights & v3.1 & en  \\
    \href{https://huggingface.co/meta-llama/Meta-Llama-3.1-70B-Instruct}{\texttt{Llama3.1-Instruct} \citep{llama3modelcard}} & 70B & Weights & v3.1 & en  \\
    \midrule
    \href{https://huggingface.co/mistralai/Mistral-7B-Instruct-v0.2}{\texttt{Mistral-Instruct} \citep{jiang2023mistral}} & 7B & Weights & v0.2 & en & Mistral AI \\
    \midrule
    \href{https://huggingface.co/01-ai/Yi-1.5-34B-Chat}{\texttt{Yi-1.5-chat} \citep{ai2024yi}} & 34B & Weights & v1.5 & zh/en & 01-AI\\
    \midrule
    \href{https://huggingface.co/internlm/internlm2_5-7b-chat}{\texttt{InternLM2.5-chat} \citep{cai2024internlm2}} & 7B & Weights & v2.5 & zh/en & \multirow{2}{*}{Shanghai AI Laboratory} \\
    \href{https://huggingface.co/internlm/internlm2_5-20b-chat}{\texttt{InternLM2.5-chat} \citep{cai2024internlm2}} & 20B & Weights & v2.5 & zh/en  \\

    \bottomrule
    \end{tabular}
    \caption{LLMs evaluated in our experiment. For LLMs with more than 70B parameters, we use SiliconCloud API.}
    \label{tab:LLMs}
\end{table*}

\subsection{Details for Evaluation Metrics}
\label{Details for evaluation Metrics}
The detailed calculation of all evaluation metrics are as follows:
\begin{equation}  
    ECE = \sum_{b=1}^{B} \frac{N_b}{N} \cdot  \left| \mathcal{A}cc(b) - \mathcal{C}onf(b) \right|, 
\end{equation}

\begin{equation}  
    MCE = \max_{b=1}^{B} \left| \mathcal{A}cc(b) - \mathcal{C}onf(b) \right|,
\end{equation}

\begin{equation}  
   ACE = \frac{1}{B} \sum_{b=1}^{B} \left| \mathcal{A}cc(b) - \mathcal{C}onf(b) \right|,
\end{equation}

\begin{equation}
    TACE(\theta) = \frac{1}{B_{\theta}} \sum_{b=1}^{B_{\theta}} \left| \mathcal{A}cc(b) - \mathcal{C}onf(b) \right|,
\end{equation}
where $B$ represents the number of probability intervals, $N_b$ represents the number of samples within the $b$-th probability interval, $N$ represents the total number of samples, $\mathcal{A}cc(b)$ represents the accuracy of the $b$-th probability interval, $\mathcal{C}onf(b)$ represents the average confidence level of the $b$-th probability interval, $\theta$ represents the confidence threshold, and $B_{\theta}$ represents the number of probability intervals composed of samples whose confidence exceeds the threshold $\theta$. In our experiments, \( N \) is set to 10, and \( \theta \) is set to 0.5.

\begin{figure*}[!t]
  \centering
  \includegraphics[height=0.3\linewidth, width=0.98\linewidth]{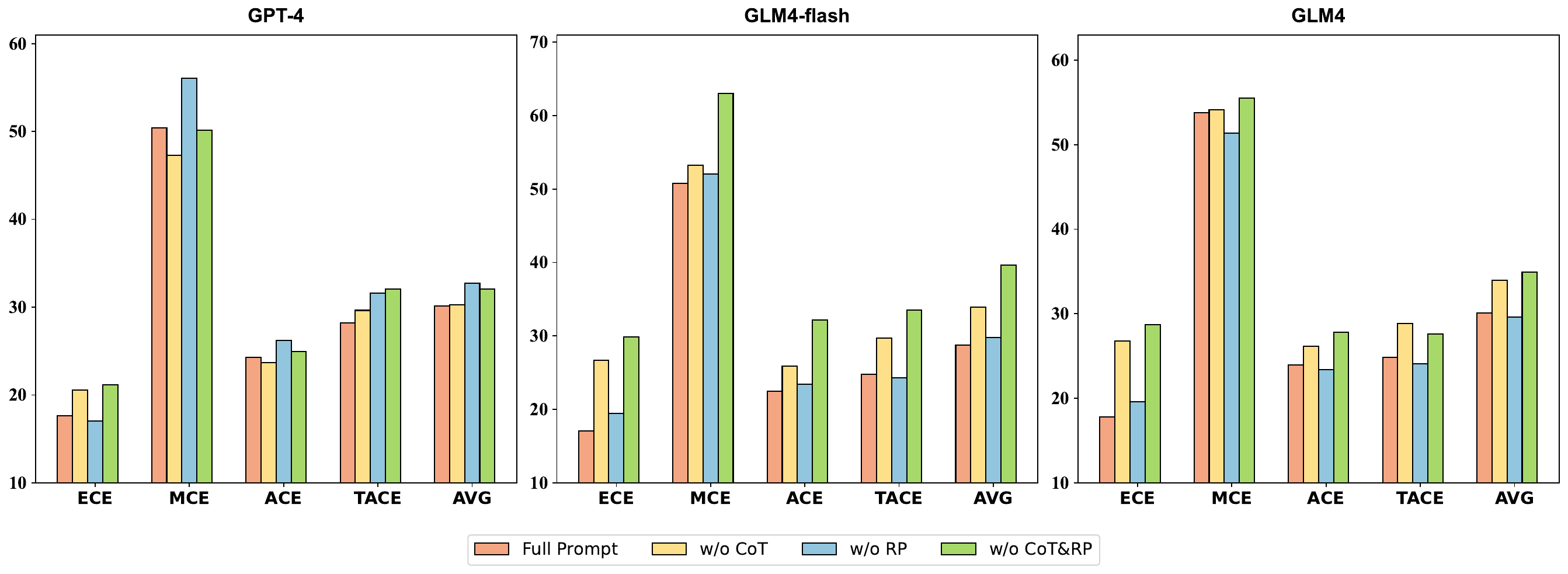}
  \caption{
  Results of experiments with GPT-4, GLM4-flash, and GLM4, studying the effects of CoT and RP prompts on LLM reliability.
  }
  \label{fig:cot}
\end{figure*}

\begin{figure*}[!t]
  \centering
  \includegraphics[width=\linewidth, height=0.3\linewidth]{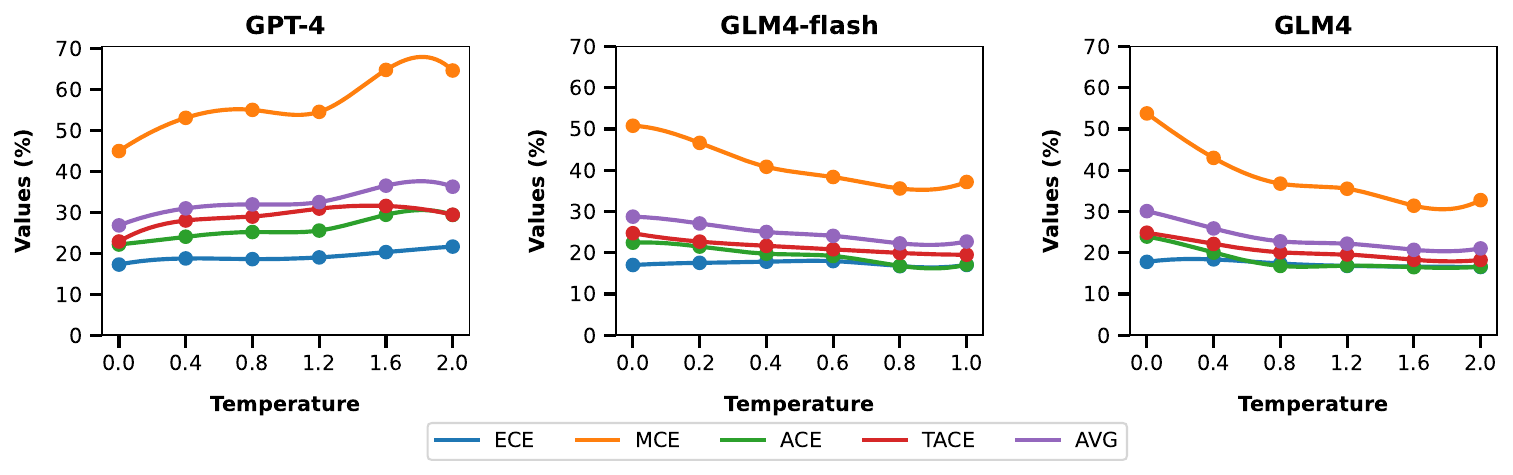}
  \caption{
    Results of experiments with GPT-4, GLM4-flash, and GLM4, studying the effects of temperature on LLM reliability.
  }
  \label{fig:tem}
\end{figure*} 

\section{Additional Experimental Results} 
\label{Further Experimental Results}

\begin{figure*}[!t]
  \centering
  \includegraphics[height=0.35\linewidth, width=0.98\linewidth]{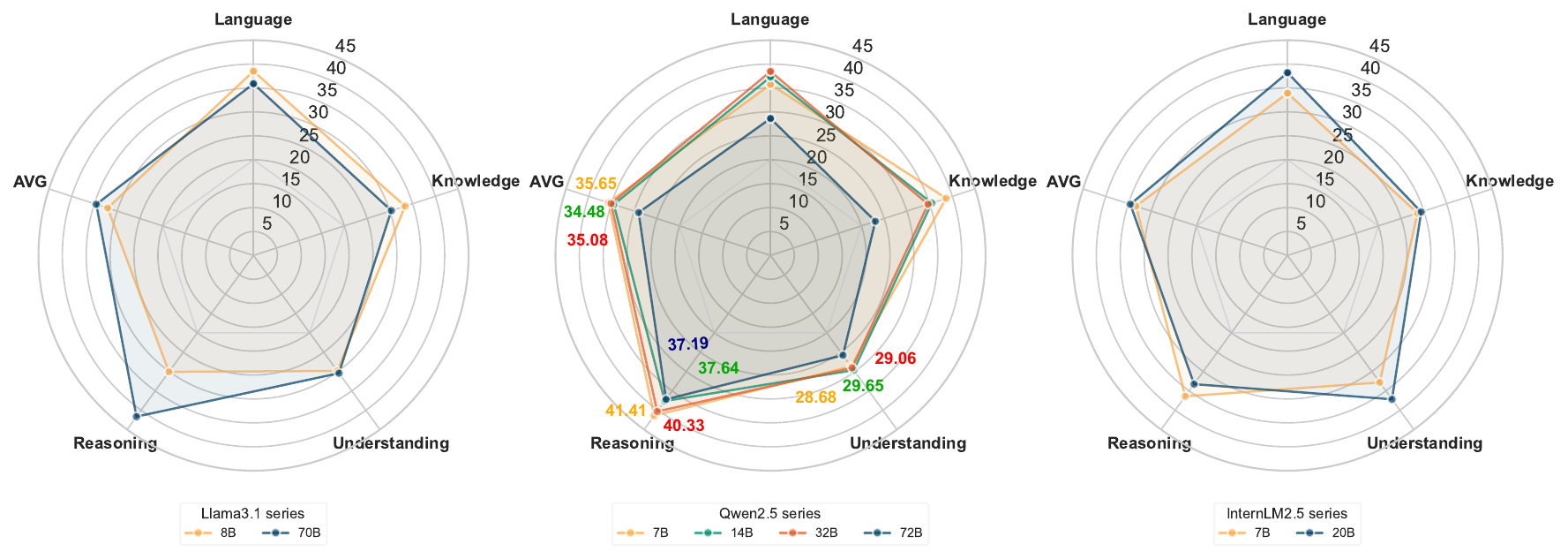}
  \caption{
Variations in \textsc{UBench} subsets for Llama3.1 series, Qwen2.5 series, and InternLM2.5 series across model sizes. Overall, as model size grows, the Llama3.1 series shows a decline in reasoning reliability but improvement in language and knowledge tasks. In contrast, the Qwen2.5 and InternLM2.5 series improve in reasoning. The InternLM2.5 series drops in language and knowledge tasks, while the Qwen2.5 series rises. For understanding tasks, the Llama3.1 series is less affected, the Qwen2.5 series improves at max size, and the InternLM2.5 series declines.
  }
  \label{fig:subset_radar}
\end{figure*} 

In this section, we present the performance of all models in each category and across all datasets.

\subsection{Performance of Different Subsets}
\begin{table*}[htbp]
\centering
\tiny
\resizebox{\textwidth}{!}{%
\begin{tabular}{
   c|l|c|c|c|c|>{\columncolor{gray!25}}c}
\toprule 
{\textbf{Category}} & \multirow{1}{*}{\textbf{LLMs}}  & \textbf{ECE (\%)} $\downarrow$ & \textbf{MCE (\%)} $\downarrow$ & \textbf{ACE (\%)} $\downarrow$ & \textbf{TACE (\%)} $\downarrow$ & \multirow{1}{*}{\textbf{AVG (\%)} $\downarrow$} \\ 
\midrule 

     \multirow{20}{*}{\includegraphics[width=0.5cm]{figs/icon/Language.png} Language} 
& \texttt{Yi-1.5-34B-Chat-16K} & 27.29 & \textcolor{magenta}{37.67} & \textcolor{blue}{22.66} & 23.55 & 27.79 \\
& \texttt{Qwen2.5-72B-Instruct} & \textcolor{blue}{19.85} & 51.98 & \textcolor{magenta}{22.00} & 20.70 & 28.63 \\
& \texttt{DeepSeek V2.5}  & 23.87 & 47.60 & 23.23 & 21.45 & 29.04 \\
& \texttt{GLM4-chat-9B} & 23.55 & 46.59 & 23.93 & 26.37 & 30.11 \\
& \texttt{Mistral-7B-Instruct-v0.2} & 21.86 & 59.08 & 24.82 & \textcolor{magenta}{18.09} & 30.96\\
& \texttt{GPT-4o} & \textcolor{magenta}{18.54} & 54.17 & 25.82 & 26.95 & 31.37 \\
& \texttt{Qwen-max} & 26.94 & \textcolor{blue}{42.61} & 28.16 & 29.70 & 31.85 \\
& \texttt{GLM4-flash} & 24.20 & 47.71 & 25.36 & 31.00 & 32.07\\
&  \texttt{GPT-4}  & 28.55 & 47.00 & 26.38 & 31.61 & 33.39 \\
& \texttt{InternLM2.5-7B-chat} & 29.04 & 58.21 & 28.40 & \textcolor{blue}{20.07} & 33.93 \\
& \texttt{Qwen2.5-7B-Instruct} & 34.36 & 49.38 & 29.13 & 30.14 & 35.75    \\
& \texttt{Llama-3.1-70B-Instruct} & 24.67 & 65.00 & 26.66 & 27.27 & 35.90 \\
& \texttt{Qwen2.5-14B-Instruct} & 35.49 & 57.50 & 27.34 & 28.99 & 37.33  \\
& \texttt{Llama-3-8B-Instruct} & 26.10 & 66.52 & 34.41 & 23.39 & 37.60 \\
& \texttt{InternLM2.5-20B-chat} & 37.50 & 52.01 & 28.14 & 35.13 & 38.20 \\
& \texttt{ChatGLM3-6B}& 28.06 & 63.36 & 29.01 & 32.93 & 38.34 \\
& \texttt{Qwen2.5-32B-Instruct} & 34.27 & 59.91 & 29.91 & 29.74 & 38.46 \\
& \texttt{Llama-3.1-8B-Instruct} & 36.02 & 59.65 & 27.24 & 31.18 & 38.52  \\
& \texttt{Qwen2-7B-Instruct} & 39.91 & 57.03 & 31.53 & 32.91 & 40.35  \\
&  \texttt{ErnieBot}& 31.44 & 70.00 & 33.90 & 28.92 & 41.06\\

    \bottomrule    
    \end{tabular}
    }

    \caption{Performance of LLMs on language subset of \textsc{UBench}.} 
    \label{language_subset}
\end{table*}

\begin{table*}[htbp]
\centering
\tiny
\resizebox{\textwidth}{!}{%
\begin{tabular}{
   c|l|c|c|c|c|>{\columncolor{gray!25}}c}
\toprule 
{\textbf{Category}} & \multirow{1}{*}{\textbf{LLMs}}  & \textbf{ECE (\%)} $\downarrow$ & \textbf{MCE (\%)} $\downarrow$ & \textbf{ACE (\%)} $\downarrow$ & \textbf{TACE (\%)} $\downarrow$ & \multirow{1}{*}{\textbf{AVG (\%)} $\downarrow$} \\ 
\midrule 

\multirow{20}{*}{\includegraphics[width=0.5cm]{figs/icon/Knowledge.png}Knowledge} 

& \texttt{Qwen-max} & 17.38 & \textcolor{magenta}{39.60} & 18.23 & \textcolor{blue}{14.69} & 22.47 \\
& \texttt{GPT-4o} & \textcolor{blue}{14.31} & 42.43 & 18.04 & 15.81 & 22.65 \\
& \texttt{Qwen2.5-72B-Instruct} & \textcolor{magenta}{12.77} & 47.27 & \textcolor{blue}{17.32} & 14.91 & 23.07 \\
& \texttt{DeepSeek V2.5}  & 25.11 & \textcolor{blue}{42.38} & \textcolor{magenta}{16.58} & \textcolor{magenta}{11.28} & 23.84 \\
& \texttt{GLM4-chat-9B} & 15.77 & 42.70 & 19.57 & 21.00 & 24.76 \\
& \texttt{GLM4-flash} & 15.33 & 44.64 & 19.98 & 23.50 & 25.86\\
& \texttt{Yi-1.5-34B-Chat-16K}  & 17.91 & 54.49 & 21.17 & 19.84 & 28.35 \\
& \texttt{InternLM2.5-7B-chat} & 27.59 & 45.52 & 21.38 & 19.93 & 28.60 \\
& \texttt{InternLM2.5-20B-chat} & 23.45 & 48.01 & 23.70 & 22.50 & 29.41 \\
& \texttt{GPT-4}  & 19.14 & 48.56 & 24.62 & 28.60 & 30.23 \\
& \texttt{Llama-3.1-70B-Instruct} & 16.99 & 52.79 & 24.16 & 27.35 & 30.32 \\
& \texttt{Mistral-7B-Instruct-v0.2} & 23.32 & 50.81 & 22.64 & 26.54 & 30.83  \\
& \texttt{Llama-3.1-8B-Instruct} & 22.78 & 61.00 & 23.94 & 25.73 & 33.36 \\
& \texttt{Qwen2.5-32B-Instruct} & 38.73 & 58.17 & 24.23 & 17.50 & 34.66 \\
& \texttt{ErnieBot}& 27.96 & 52.24 & 29.20 & 31.97 & 35.34 \\
& \texttt{Qwen2.5-14B-Instruct} & 38.46 & 55.15 & 27.09 & 21.22 & 35.48 \\
& \texttt{Llama-3-8B-Instruct} & 26.18 & 64.00 & 25.72 & 28.40 & 36.07  \\
& \texttt{Qwen2.5-7B-Instruct} & 38.68 & 63.00 & 28.86 & 24.02 & 38.64 \\
& \texttt{Qwen2-7B-Instruct} & 39.86 & 55.10 & 30.40 & 30.07 & 38.86 \\
& \texttt{ChatGLM3-6B}& 24.85 & 68.65 & 31.07 & 41.38 & 41.49 \\

\bottomrule    
\end{tabular}
    }

    \caption{Performance of LLMs on knowledge subset of \textsc{UBench}.} 
    \label{knowledge_subset}
\end{table*}

\begin{table*}[htbp]
\centering
\tiny
\resizebox{\textwidth}{!}{%
\begin{tabular}{
   c|l|c|c|c|c|>{\columncolor{gray!25}}c}
\toprule 
{\textbf{Category}} & \multirow{1}{*}{\textbf{LLMs}}  & \textbf{ECE (\%)} $\downarrow$ & \textbf{MCE (\%)} $\downarrow$ & \textbf{ACE (\%)} $\downarrow$ & \textbf{TACE (\%)} $\downarrow$ & \multirow{1}{*}{\textbf{AVG (\%)} $\downarrow$} \\ 
\midrule

    \multirow{20}{*}{\includegraphics[width=0.55cm]{figs/icon/Understanding.png}Understanding}
    
& \texttt{Qwen-max} & 14.89 & \textcolor{magenta}{37.25} & \textcolor{blue}{17.21} & \textcolor{magenta}{14.29} & 20.91 \\
&  \texttt{ErnieBot}& 15.15 & \textcolor{blue}{40.31} & 17.64 & 18.65 & 22.94\\
& \texttt{DeepSeek V2.5}  & 21.29 & 45.18 & \textcolor{magenta}{17.13} & \textcolor{blue}{14.92} & 24.63 \\
& \texttt{GPT-4o} & 17.44 & 43.41 & 19.70 & 18.36 & 24.73 \\
& \texttt{Yi-1.5-34B-Chat-16K}  & \textcolor{blue}{14.68} & 50.93 & 19.06 & 15.37 & 25.01\\
& \texttt{Qwen2.5-72B-Instruct} & \textcolor{magenta}{14.54} & 48.11 & 20.59 & 19.90 & 25.79 \\
& \texttt{Llama-3-8B-Instruct} & 15.54 & 47.92 & 21.93 & 20.85 & 26.56\\
& \texttt{GLM4-flash} & 16.26 & 52.22 & 21.90 & 19.36 & 27.44\\
& \texttt{Qwen2.5-7B-Instruct} & 27.83 & 48.48 & 20.43 & 17.99 & 28.68    \\
& \texttt{Qwen2.5-32B-Instruct} & 27.95 & 46.81 & 22.52 & 18.98 & 29.06 \\
& \texttt{GLM4-chat-9B} & 15.18 & 55.10 & 24.97 & 21.51 & 29.19 \\
& \texttt{Qwen2.5-14B-Instruct} & 27.88 & 49.76 & 22.94 & 18.03 & 29.65  \\
& \texttt{Llama-3.1-8B-Instruct} & 21.21 & 52.19 & 23.27 & 22.70 & 29.84  \\
& \texttt{Llama-3.1-70B-Instruct} & 22.36 & 51.48 & 23.38 & 24.48 & 30.43  \\
& \texttt{InternLM2.5-7B-chat} & 25.76 & 51.89 & 26.79 & 25.99 & 32.82 \\
&  \texttt{GPT-4} & 17.12 & 63.75 & 27.30 & 28.61 & 34.20\\
& \texttt{Qwen2-7B-Instruct} & 30.96 & 55.49 & 25.74 & 25.60 & 34.45  \\
& \texttt{InternLM2.5-20B-chat} & 21.25 & 67.76 & 26.48 & 33.23 & 37.18 \\
& \texttt{Mistral-7B-Instruct-v0.2} & 24.79 & 66.86 & 32.28 & 34.32 & 39.56 \\
& \texttt{ChatGLM3-6B}& 26.81 & 67.21 & 30.19 & 35.12 & 39.83 \\

    \bottomrule    
    \end{tabular}%
    }

    \caption{Performance of LLMs on understanding subset of \textsc{UBench}.} 
    \label{understanding_subset}
\end{table*}

\begin{table*}[htbp]
\centering
\tiny
\resizebox{\textwidth}{!}{%
\begin{tabular}{
   c|l|c|c|c|c|>{\columncolor{gray!25}}c}
\toprule 
{\textbf{Category}} & \multirow{1}{*}{\textbf{LLMs}}  & \textbf{ECE (\%)} $\downarrow$ & \textbf{MCE (\%)} $\downarrow$ & \textbf{ACE (\%)} $\downarrow$ & \textbf{TACE (\%)} $\downarrow$ & \multirow{1}{*}{\textbf{AVG (\%)} $\downarrow$} \\ 
\midrule 

    \multirow{20}{*}{\includegraphics[width=0.5cm]{figs/icon/Reasoning.png}Reasoning}

& \texttt{Llama-3-8B-Instruct} & 16.01 & \textcolor{blue}{42.78} & \textcolor{magenta}{16.33} & \textcolor{magenta}{16.39} & 22.88\\
&  \texttt{GPT-4}  & 10.94 & \textcolor{magenta}{38.49} & \textcolor{blue}{19.27} & \textcolor{blue}{25.53} & 23.56 \\
& \texttt{GPT-4o} & \textcolor{magenta}{6.01} & 52.86 & 21.51 & 25.66 & 26.51 \\
& \texttt{Yi-1.5-34B-Chat-16K}  & 22.56 & 45.10 & 20.18 & 25.68 & 28.38  \\
& \texttt{Llama-3.1-8B-Instruct} & 11.56 & 56.25 & 23.24 & 29.30 & 30.09  \\
& \texttt{GLM4-flash} & 15.04 & 55.32 & 23.15 & 28.18 & 30.42\\
& \texttt{Qwen-max} & 18.47 & 59.29 & 24.35 & 30.85 & 33.24 \\
& \texttt{InternLM2.5-20B-chat} & 13.68 & 60.92 & 26.04 & 32.41 & 33.26 \\
& \texttt{DeepSeek V2.5}  & 14.85 & 63.90 & 27.81 & 32.56 & 34.78  \\
& \texttt{GLM4-chat-9B} & 18.92 & 64.29 & 25.80 & 30.59 & 34.90 \\
& \texttt{Mistral-7B-Instruct-v0.2} & 17.98 & 69.17 & 25.57 & 27.97 & 35.17\\
& \texttt{ErnieBot} & 24.36 & 57.97 & 28.96 & 32.99 & 36.07 \\
& \texttt{InternLM2.5-7B-chat} & 23.73 & 64.73 & 27.06 & 30.13 & 36.41 \\
& \texttt{Qwen2.5-72B-Instruct} & \textcolor{blue}{7.51} & 68.93 & 31.01 & 41.31 & 37.19 \\
& \texttt{Qwen2.5-14B-Instruct} & 14.86 & 67.38 & 30.46 & 37.88 & 37.64  \\
& \texttt{Qwen2.5-32B-Instruct} & 14.90 & 71.99 & 32.50 & 41.92 & 40.33\\
& \texttt{Qwen2.5-7B-Instruct} & 15.07 & 76.43 & 32.60 & 41.55 & 41.41    \\
& \texttt{Llama-3.1-70B-Instruct} & 11.86 & 73.70 & 33.48 & 47.56 & 41.65  \\
& \texttt{ChatGLM3-6B}& 22.08 & 78.81 & 30.04 & 38.52 & 42.36 \\
& \texttt{Qwen2-7B-Instruct} & 39.57 & 70.24 & 31.91 & 27.81 & 42.38 \\

   \bottomrule    
    \end{tabular}%
    }

    \caption{Performance of LLMs on reasoning subset of \textsc{UBench}.} 
    \label{reasoning_subset}
\end{table*}

The performance of LLMs on \textsc{UBench} for the four subsets of knowledge, language, understanding, and reasoning is shown in Table~\cref{language_subset,knowledge_subset,understanding_subset,reasoning_subset}. As shown in Figure~\ref{fig:subset_radar}, the changes in model size and the enhancement of four key capabilities are linked to different model series. Generally, as model size increases, the Llama3.1 series shows a decrease in reasoning reliability but an improvement in language and knowledge tasks. In contrast, the Qwen2.5 and InternLM2.5 series progress in reasoning. The language and knowledge performance of the InternLM2.5 series declines, while that of the Qwen2.5 series rises. For understanding tasks, the Llama3.1 series is less affected. The Qwen2.5 series improves most at the maximum size, and the InternLM2.5 series declines.

\paragraph{Performance of Language Subset.} 
In this category, Yi-1.5-34B and Qwen2.5-72B   perform similarly in this dataset and are considered outstanding, which indicates their high reliability in language tasks. In comparison, GLM4-chat-9B demonstrates an 8.23\% improvement over the previous generation ChatGLM3-6B, while Qwen2.5-7B-Instruct shows a 4.6\% increase compared to Qwen2-7B-Instruct. However, we also find that the performance of Llama-3.1-8B-Instruct is lower than that of Llama-3-8B-Instruct. This seemingly implies that enhancing the model's other capabilities (as model upgrades often lead to improvements in certain areas) does not necessarily result in an increase in the model's reliability. This emphasizes the importance of incorporating reliability assessments into model evaluations.

\paragraph{Performance of Knowledge Subset.} In knowledge-based datasets, closed-source models generally outperform open-source models. Specifically, Qwen-max is the best-performing model, but the gap between it and the second-place GPT-4o and the third-place Qwen2.5-72B-Instruct is minimal. Although GPT-4o and Qwen2.5-72B-Instruct have lower ECE values, their higher MCE negatively affects their rankings. This once again demonstrates the superiority and comprehensiveness of the multiple metrics in our benchmark. GLM4-chat-9B, as a 10B-sized model, outperforms several larger models, particularly achieving 5.56\% better performance than Llama-3.1-70B-Instruct. Similarly, InternLM2.5-7B-chat also performs well, narrowly surpassing InternLM2.5-20B-chat, indicating that smaller models can also exhibit good reliability in knowledge-based tasks. Nevertheless, the performance of the Qwen2.5 series improves as the model size increases, demonstrating its unique superiority. These seem to suggest that in knowledge-based tasks, the model size is not the key factor influencing the model's reliability.

\paragraph{Performance of Understanding Subset.} In this dataset, Qwen-max performs the best, followed by ErnieBot and DeepSeek V2.5. Closed-source models, such as Qwen-max, ErnieBot, and GPT-4o, outperform most open-source models, indicating that they exhibit higher reliability in understanding tasks. Small-scale models like Llama-3-8B-Instruct and Qwen2.5-7B-Instruct still demonstrate strong competitiveness, with Llama-3-8B-Instruct performing only 0.77\% worse than Qwen2.5-72B-Instruct. The larger model, InternLM2.5-20B, shows less favorable performance, trailing behind many models with fewer parameters. This once again suggests that the size of the model is not the only influencing factor when it comes to model reliability.

\paragraph{Performance of Reasoning Subset.} In general, Llama-3-8B-Instruct, GPT-4, and GPT-4o rank in the top three, with a significant gap between the first two and the latter. Although GPT-4o has the lowest ECE value, its MCE severely impacts its overall performance. Similar to Llama-3-8B-Instruct, Llama-3.1-8B-Instruct also demonstrates outstanding performance, outperforming several closed-source models, including GLM4-flash and Qwen-max, and even some larger-scale models such as InternLM2.5-20B, Qwen2.5-\{14, 32, 72\}B, and Llama-3.1-70B-Instruct. Despite excelling in other three categories, DeepSeek V2.5 does not stand out in this task, with performance comparable to GLM4-chat-9B. Models like ChatGLM3-6B and Qwen2-7B-Instruct fall behind compared to the most advanced models in their series. The high MCE values of Llama-3.1-70B-Instruct and Qwen2.5-\{14, 32, 72\}B-Instruct significantly affect their rankings, raising concerns about model reliability under extreme conditions.

\begin{table}[]
\scriptsize
\centering
\begin{tabular}{
   l|c|c|c|c|>{\columncolor{gray!25}}c}
\toprule 
\multirow{1}{*}{\textbf{Dataset}}  & \textbf{ECE}       & \textbf{MCE}       & \textbf{ACE}  & \textbf{TACE}   &\multirow{1}{*}{\textbf{AVG}}
                              
                               \\ \midrule

\texttt{WIC} & 19.54 & 75.00 & 31.62 & 32.71 & 39.72 \\
\texttt{WSC} & 17.36 & 55.00 & 21.21 & 21.24 & 28.70 \\
\texttt{COLA} & 25.18 & 45.00 & 26.99 & 24.25 & 30.36 \\
\texttt{QQP} & 12.07 & 41.67 & 23.48 & 29.59 & 26.70 \\
\texttt{CommonSenseQA} & 4.60 & 35.00 & 8.82 & 5.74 & 13.54 \\
\texttt{TruthfulQA} & 11.96 & 35.00 & 17.18 & 11.81 & 18.99 \\
\texttt{popqa-parents-lying} & 25.78 & 85.00 & 31.59 & 33.46 & 43.96 \\
\texttt{NQ} & 15.48 & 32.14 & 21.02 & 19.09 & 21.93 \\
\texttt{WikiQA} & 13.73 & 25.00 & 11.58 & 8.95 & 14.82 \\
\texttt{RACE} & 14.34 & 25.00 & 15.62 & 14.73 & 17.42 \\
\texttt{MMLU} & 7.15 & 40.71 & 10.53 & 6.92 & 16.33 \\
\texttt{LogiQA} & 15.94 & 22.14 & 13.80 & 16.69 & 17.14 \\
\texttt{SWAG} & 25.64 & 42.78 & 23.62 & 18.29 & 27.58 \\
\texttt{Cosmos QA} & 42.11 & 45.00 & 26.44 & 22.50 & 34.01 \\
\texttt{Social IQA} & 13.28 & 31.67 & 16.33 & 9.90 & 17.79 \\
\texttt{DialogSum} & 16.12 & 65.00 & 32.56 & 35.24 & 37.23 \\
\texttt{CommonGen} & 4.96 & 75.00 & 18.70 & 22.57 & 30.31 \\
\texttt{GSM8K} & 2.88 & 25.00 & 10.22 & 4.18 & 10.57 \\
\texttt{AQuA} & 9.88 & 65.00 & 25.53 & 18.12 & 29.63 \\
\texttt{AddSub} & 5.22 & 85.00 & 28.60 & 41.14 & 39.99 \\
\texttt{MultiArith} & 4.42 & 15.00 & 7.94 & 10.00 & 9.34 \\
\texttt{SingleEq} & 5.34 & 85.00 & 36.67 & 55.00 & 45.50 \\
\texttt{SVAMP} & 3.52 & 65.00 & 30.31 & 41.47 & 35.08 \\
\texttt{PIQA} & 10.81 & 30.00 & 11.27 & 9.68 & 15.44 \\

    \bottomrule
\end{tabular}
 \caption{The performance of the \texttt{GPT-4o-2024-08-06} model across multiple datasets.}
\label{model-gpt-4o-2024-08-06}
\end{table}

\begin{table}[]
\scriptsize
\centering
\begin{tabular}{
   l|c|c|c|c|>{\columncolor{gray!25}}c}
\toprule 
\multirow{1}{*}{\textbf{Dataset}}  & \textbf{ECE}       & \textbf{MCE}       & \textbf{ACE}  & \textbf{TACE}   &\multirow{1}{*}{\textbf{AVG}}
                              
                               \\ \midrule

\texttt{WIC} & 29.50 & 35.40 & 19.77 & 35.40 & 30.02 \\
\texttt{WSC} & 30.65 & 75.00 & 39.91 & 34.63 & 45.05 \\
\texttt{COLA} & 34.25 & 45.00 & 30.77 & 38.53 & 37.14 \\
\texttt{QQP} & 19.80 & 32.62 & 15.06 & 17.88 & 21.34 \\
\texttt{CommonSenseQA} & 11.65 & 75.00 & 22.84 & 27.87 & 34.34 \\
\texttt{TruthfulQA} & 11.00 & 65.00 & 32.50 & 40.87 & 37.34 \\
\texttt{popqa-parents-lying} & 33.00 & 36.67 & 30.12 & 36.67 & 34.12 \\
\texttt{NQ} & 22.20 & 31.15 & 19.82 & 20.84 & 23.50 \\
\texttt{WikiQA} & 17.85 & 35.00 & 17.84 & 16.74 & 21.86 \\
\texttt{RACE} & 9.30 & 31.67 & 18.21 & 20.95 & 20.03 \\
\texttt{MMLU} & 22.35 & 75.00 & 34.01 & 39.66 & 42.75 \\
\texttt{LogiQA} & 25.90 & 65.00 & 29.00 & 40.33 & 40.06 \\
\texttt{SWAG} & 21.05 & 85.00 & 32.80 & 23.75 & 40.65 \\
\texttt{Cosmos QA} & 11.07 & 85.00 & 24.86 & 18.59 & 34.88 \\
\texttt{Social IQA} & 16.35 & 35.00 & 19.58 & 21.84 & 23.19 \\
\texttt{DialogSum} & 22.35 & 68.33 & 33.91 & 31.28 & 38.97 \\
\texttt{CommonGen} & 8.60 & 65.00 & 26.00 & 32.50 & 33.02 \\
\texttt{GSM8K} & 10.00 & 15.34 & 8.83 & 15.34 & 12.38 \\
\texttt{AQuA} & 16.06 & 35.00 & 18.05 & 24.15 & 23.31 \\
\texttt{AddSub} & 13.00 & 85.00 & 35.21 & 51.72 & 46.23 \\
\texttt{MultiArith} & 5.40 & 6.61 & 5.24 & 6.61 & 5.96 \\
\texttt{SingleEq} & 5.40 & 7.50 & 5.11 & 7.50 & 6.38 \\
\texttt{SVAMP} & 6.50 & 85.00 & 43.01 & 55.75 & 47.56 \\
\texttt{PIQA} & 20.25 & 35.00 & 19.43 & 17.62 & 23.08 \\
    \bottomrule
\end{tabular}
 \caption{The performance of the  \texttt{GPT-4} model across multiple datasets.}
\label{model-gpt4}
\end{table}

\begin{table}[]
\scriptsize
\centering
\begin{tabular}{
   l|c|c|c|c|>{\columncolor{gray!25}}c}
\toprule 
\multirow{1}{*}{\textbf{Dataset}}  & \textbf{ECE}       & \textbf{MCE}       & \textbf{ACE}  & \textbf{TACE}   &\multirow{1}{*}{\textbf{AVG}}
                              
                               \\ \midrule

\texttt{WIC} & 35.15 & 75.00 & 38.10 & 21.30 & 42.39 \\
\texttt{WSC} & 37.95 & 55.00 & 34.13 & 41.22 & 42.08 \\
\texttt{COLA} & 20.90 & 75.00 & 31.54 & 14.75 & 35.55 \\
\texttt{QQP} & 31.75 & 75.00 & 31.82 & 38.41 & 44.24 \\
\texttt{CommonSenseQA} & 19.70 & 35.00 & 18.60 & 14.78 & 22.02 \\
\texttt{TruthfulQA} & 31.05 & 55.00 & 26.18 & 21.41 & 33.41 \\
\texttt{popqa-parents-lying} & 38.50 & 44.53 & 31.51 & 33.26 & 36.95 \\
\texttt{NQ} & 33.60 & 85.00 & 45.53 & 66.80 & 57.73 \\
\texttt{WikiQA} & 16.95 & 41.67 & 24.20 & 23.61 & 26.61 \\
\texttt{RACE} & 14.45 & 28.33 & 13.50 & 16.67 & 18.24 \\
\texttt{MMLU} & 24.70 & 75.00 & 27.45 & 26.58 & 38.43 \\
\texttt{LogiQA} & 21.45 & 30.29 & 20.07 & 21.46 & 23.32 \\
\texttt{SWAG} & 12.70 & 35.00 & 14.13 & 17.38 & 19.80 \\
\texttt{Cosmos QA} & 10.28 & 35.00 & 13.47 & 10.00 & 17.19 \\
\texttt{Social IQA} & 10.20 & 33.89 & 12.98 & 16.75 & 18.46 \\
\texttt{DialogSum} & 18.15 & 50.00 & 25.05 & 23.33 & 29.13 \\
\texttt{CommonGen} & 9.30 & 35.00 & 14.50 & 17.06 & 18.96 \\
\texttt{GSM8K} & 21.85 & 45.00 & 30.25 & 28.62 & 31.43 \\
\texttt{AQuA} & 40.00 & 72.44 & 35.87 & 32.96 & 45.32 \\
\texttt{AddSub} & 18.90 & 45.00 & 24.43 & 21.20 & 27.38 \\
\texttt{MultiArith} & 41.80 & 75.00 & 42.41 & 64.17 & 55.84 \\
\texttt{SingleEq} & 15.96 & 48.33 & 20.73 & 27.89 & 28.23 \\
\texttt{SVAMP} & 16.66 & 85.00 & 32.45 & 41.60 & 43.93 \\
\texttt{PIQA} & 15.35 & 35.00 & 16.61 & 14.46 & 20.36 \\
    \bottomrule
\end{tabular}
 \caption{The performance of the  \texttt{ERNIE-Bot-4.0} model across multiple datasets.}
\label{model-ERNIE-Bot-4.0}
\end{table}

\begin{table}[]
\scriptsize
\centering
\begin{tabular}{
   l|c|c|c|c|>{\columncolor{gray!25}}c}
\toprule 
\multirow{1}{*}{\textbf{Dataset}}  & \textbf{ECE}       & \textbf{MCE}       & \textbf{ACE}  & \textbf{TACE}   &\multirow{1}{*}{\textbf{AVG}}
                              
                               \\ \midrule

\texttt{WIC} & 13.84 & 29.17 & 14.97 & 17.02 & 18.75 \\
\texttt{WSC} & 37.28 & 85.00 & 35.31 & 26.58 & 46.04 \\
\texttt{COLA} & 25.68 & 35.00 & 22.83 & 21.98 & 26.37 \\
\texttt{QQP} & 18.68 & 41.25 & 19.83 & 20.24 & 25.00 \\
\texttt{CommonSenseQA} & 32.71 & 38.15 & 16.16 & 3.87 & 22.72 \\
\texttt{TruthfulQA} & 24.61 & 33.66 & 12.58 & 9.10 & 19.99 \\
\texttt{popqa-parents-lying} & 26.74 & 75.00 & 27.30 & 23.37 & 38.10 \\
\texttt{NQ} & 20.69 & 31.67 & 16.95 & 11.36 & 20.17 \\
\texttt{WikiQA} & 20.82 & 33.40 & 9.93 & 8.69 & 18.21 \\
\texttt{RACE} & 13.27 & 20.00 & 13.17 & 12.92 & 14.84 \\
\texttt{MMLU} & 23.86 & 34.84 & 13.11 & 6.78 & 19.65 \\
\texttt{LogiQA} & 37.01 & 85.00 & 27.16 & 20.16 & 42.33 \\
\texttt{SWAG} & 31.62 & 40.17 & 13.91 & 12.19 & 24.47 \\
\texttt{Cosmos QA} & 24.08 & 65.00 & 23.57 & 18.10 & 32.69 \\
\texttt{Social IQA} & 26.26 & 36.40 & 16.35 & 8.68 & 21.92 \\
\texttt{DialogSum} & 8.84 & 45.00 & 15.36 & 19.98 & 22.30 \\
\texttt{CommonGen} & 5.38 & 35.00 & 14.44 & 20.56 & 18.84 \\
\texttt{GSM8K} & 4.76 & 85.00 & 35.53 & 49.06 & 43.59 \\
\texttt{AQuA} & 11.79 & 50.00 & 22.22 & 21.20 & 26.30 \\
\texttt{AddSub} & 22.43 & 85.00 & 42.36 & 56.72 & 51.63 \\
\texttt{MultiArith} & 8.20 & 35.00 & 18.24 & 5.00 & 16.61 \\
\texttt{SingleEq} & 18.93 & 85.00 & 37.79 & 48.21 & 47.48 \\
\texttt{SVAMP} & 18.60 & 75.00 & 30.38 & 40.74 & 41.18 \\
\texttt{PIQA} & 19.23 & 32.33 & 8.12 & 6.99 & 16.67 \\
    \bottomrule
\end{tabular}
 \caption{The performance of the  \texttt{Deepseek-chat} model across multiple datasets.}
\label{model-deepseek-chat}
\end{table}

\begin{table}[]
\scriptsize
\centering
\begin{tabular}{
   l|c|c|c|c|>{\columncolor{gray!25}}c}
\toprule 
\multirow{1}{*}{\textbf{Dataset}}  & \textbf{ECE}       & \textbf{MCE}       & \textbf{ACE}  & \textbf{TACE}   &\multirow{1}{*}{\textbf{AVG}}
                              
                               \\ \midrule

\texttt{WIC} & 23.74 & 39.06 & 23.32 & 24.08 & 27.55 \\
\texttt{WSC} & 45.92 & 85.00 & 62.06 & 60.11 & 63.27 \\
\texttt{COLA} & 13.46 & 14.34 & 10.91 & 11.11 & 12.45 \\
\texttt{QQP} & 24.62 & 32.05 & 16.35 & 23.49 & 24.13 \\
\texttt{CommonSenseQA} & 5.76 & 15.00 & 7.79 & 8.72 & 9.32 \\
\texttt{TruthfulQA} & 15.98 & 40.00 & 17.48 & 14.09 & 21.89 \\
\texttt{popqa-parents-lying} & 34.90 & 75.00 & 33.79 & 18.75 & 40.61 \\
\texttt{NQ} & 17.70 & 33.00 & 18.05 & 14.06 & 20.70 \\
\texttt{WikiQA} & 12.56 & 35.00 & 14.02 & 17.84 & 19.86 \\
\texttt{RACE} & 8.74 & 18.33 & 8.72 & 6.68 & 10.62 \\
\texttt{MMLU} & 14.88 & 25.00 & 12.65 & 9.72 & 15.56 \\
\texttt{LogiQA} & 16.21 & 35.22 & 12.95 & 10.85 & 18.81 \\
\texttt{SWAG} & 20.82 & 43.24 & 24.31 & 17.37 & 26.44 \\
\texttt{Cosmos QA} & 10.33 & 33.89 & 14.26 & 16.40 & 18.72 \\
\texttt{Social IQA} & 15.82 & 47.35 & 22.07 & 20.09 & 26.33 \\
\texttt{DialogSum} & 25.30 & 65.00 & 28.54 & 20.53 & 34.84 \\
\texttt{CommonGen} & 7.02 & 30.00 & 14.16 & 12.69 & 15.97 \\
\texttt{GSM8K} & 14.74 & 49.29 & 23.31 & 19.49 & 26.71 \\
\texttt{AQuA} & 23.17 & 46.82 & 25.17 & 26.41 & 30.39 \\
\texttt{AddSub} & 23.45 & 85.00 & 28.99 & 45.67 & 45.78 \\
\texttt{MultiArith} & 16.49 & 65.00 & 29.95 & 36.11 & 36.89 \\
\texttt{SingleEq} & 21.21 & 77.86 & 24.64 & 39.71 & 40.85 \\
\texttt{SVAMP} & 18.65 & 72.50 & 26.59 & 35.22 & 38.24 \\
\texttt{PIQA} & 11.56 & 18.58 & 11.81 & 13.35 & 13.82 \\
    \bottomrule
\end{tabular}
 \caption{The performance of the  \texttt{Qwen-max} model across multiple datasets.}
\label{model-qwen-max}
\end{table}

\begin{table}[]
\scriptsize
\centering
\begin{tabular}{
   l|c|c|c|c|>{\columncolor{gray!25}}c}
\toprule 
\multirow{1}{*}{\textbf{Dataset}}  & \textbf{ECE}       & \textbf{MCE}       & \textbf{ACE}  & \textbf{TACE}   &\multirow{1}{*}{\textbf{AVG}}
                              
                               \\ \midrule

\texttt{WIC} & 44.14 & 75.00 & 43.18 & 45.21 & 51.88 \\
\texttt{WSC} & 44.28 & 47.31 & 32.35 & 33.92 & 39.47 \\
\texttt{COLA} & 35.08 & 54.15 & 20.90 & 20.80 & 32.73 \\
\texttt{QQP} & 36.12 & 51.67 & 29.70 & 31.73 & 37.31 \\
\texttt{CommonSenseQA} & 36.08 & 55.95 & 31.37 & 26.87 & 37.57 \\
\texttt{TruthfulQA} & 45.68 & 48.53 & 27.26 & 23.47 & 36.23 \\
\texttt{popqa-parents-lying} & 42.26 & 43.76 & 35.70 & 43.76 & 41.37 \\
\texttt{NQ} & 37.26 & 85.00 & 36.75 & 33.61 & 48.16 \\
\texttt{WikiQA} & 38.00 & 42.24 & 20.93 & 22.63 & 30.95 \\
\texttt{RACE} & 35.64 & 65.00 & 27.09 & 26.96 & 38.67 \\
\texttt{MMLU} & 38.32 & 44.10 & 22.57 & 23.52 & 32.13 \\
\texttt{LogiQA} & 38.48 & 55.00 & 29.92 & 26.85 & 37.56 \\
\texttt{SWAG} & 35.38 & 75.00 & 30.50 & 23.13 & 41.00 \\
\texttt{Cosmos QA} & 28.95 & 41.10 & 21.39 & 26.86 & 29.57 \\
\texttt{Social IQA} & 30.52 & 75.00 & 29.70 & 22.70 & 39.48 \\
\texttt{DialogSum} & 10.78 & 45.00 & 21.67 & 21.67 & 24.78 \\
\texttt{CommonGen} & 29.62 & 43.68 & 23.07 & 33.08 & 32.36 \\
\texttt{GSM8K} & 39.32 & 95.00 & 38.61 & 25.95 & 49.72 \\
\texttt{AQuA} & 39.28 & 58.33 & 26.80 & 30.36 & 38.69 \\
\texttt{AddSub} & 41.10 & 61.67 & 32.25 & 32.23 & 41.81 \\
\texttt{MultiArith} & 37.00 & 95.00 & 32.57 & 25.97 & 47.63 \\
\texttt{SingleEq} & 40.46 & 51.67 & 32.41 & 32.73 & 39.32 \\
\texttt{SVAMP} & 38.02 & 45.00 & 23.41 & 26.14 & 33.14 \\
\texttt{PIQA} & 41.84 & 85.00 & 37.34 & 21.31 & 46.37 \\

    \bottomrule
\end{tabular}
 \caption{The performance of the  \texttt{Qwen2-7B-Instruct} model across multiple datasets.}
\label{model-Qwen2-7B-Instruct}
\end{table}

\begin{table}[]
\scriptsize
\centering
\begin{tabular}{
   l|c|c|c|c|>{\columncolor{gray!25}}c}
\toprule 
\multirow{1}{*}{\textbf{Dataset}}  & \textbf{ECE}       & \textbf{MCE}       & \textbf{ACE}  & \textbf{TACE}   &\multirow{1}{*}{\textbf{AVG}}
                              
                               \\ \midrule

\texttt{WIC} & 22.90 & 35.82 & 15.77 & 17.70 & 23.05 \\
\texttt{WSC} & 48.54 & 85.00 & 52.34 & 54.22 & 60.02 \\
\texttt{COLA} & 41.97 & 45.00 & 30.11 & 27.84 & 36.23 \\
\texttt{QQP} & 24.03 & 31.71 & 18.31 & 20.78 & 23.71 \\
\texttt{CommonSenseQA} & 44.12 & 65.00 & 33.74 & 28.75 & 42.90 \\
\texttt{TruthfulQA} & 40.86 & 65.00 & 23.43 & 16.91 & 36.55 \\
\texttt{popqa-parents-lying} & 35.60 & 55.00 & 30.45 & 33.27 & 38.58 \\
\texttt{NQ} & 34.82 & 65.00 & 25.66 & 22.54 & 37.00 \\
\texttt{WikiQA} & 37.98 & 65.00 & 31.02 & 18.62 & 38.16 \\
\texttt{RACE} & 25.44 & 75.00 & 21.18 & 13.75 & 33.84 \\
\texttt{MMLU} & 32.15 & 38.03 & 17.84 & 18.83 & 26.71 \\
\texttt{LogiQA} & 33.56 & 41.60 & 21.86 & 24.56 & 30.39 \\
\texttt{SWAG} & 23.22 & 34.05 & 9.51 & 5.38 & 18.04 \\
\texttt{Cosmos QA} & 43.81 & 85.00 & 39.18 & 23.83 & 47.96 \\
\texttt{Social IQA} & 26.37 & 33.11 & 16.10 & 10.66 & 21.56 \\
\texttt{DialogSum} & 28.72 & 38.57 & 23.51 & 27.17 & 29.49 \\
\texttt{CommonGen} & 9.34 & 42.50 & 14.27 & 19.72 & 21.46 \\
\texttt{GSM8K} & 8.03 & 65.00 & 27.17 & 30.35 & 32.64 \\
\texttt{AQuA} & 17.40 & 45.00 & 21.92 & 27.36 & 27.92 \\
\texttt{AddSub} & 9.13 & 85.00 & 40.28 & 57.66 & 48.02 \\
\texttt{MultiArith} & 7.28 & 85.00 & 28.03 & 38.11 & 39.61 \\
\texttt{SingleEq} & 9.66 & 85.00 & 38.92 & 55.15 & 47.18 \\
\texttt{SVAMP} & 11.24 & 85.00 & 32.81 & 51.05 & 45.03 \\
\texttt{PIQA} & 42.72 & 85.00 & 39.07 & 31.20 & 49.50 \\
    \bottomrule
\end{tabular}
 \caption{The performance of the  \texttt{Qwen2.5-7B-Instruct} model across multiple datasets.}
\label{model-Qwen2.5-7B-Instruct}
\end{table}

\begin{table}[]
\scriptsize
\centering
\begin{tabular}{
   l|c|c|c|c|>{\columncolor{gray!25}}c}
\toprule 
\multirow{1}{*}{\textbf{Dataset}}  & \textbf{ECE}       & \textbf{MCE}       & \textbf{ACE}  & \textbf{TACE}   &\multirow{1}{*}{\textbf{AVG}}
                              
                               \\ \midrule

\texttt{WIC} & 22.92 & 45.00 & 24.88 & 21.31 & 28.53 \\
\texttt{WSC} & 52.57 & 85.00 & 45.48 & 45.30 & 57.09 \\
\texttt{COLA} & 40.99 & 65.00 & 23.74 & 26.93 & 39.17 \\
\texttt{QQP} & 25.47 & 35.00 & 15.25 & 22.41 & 24.53 \\
\texttt{CommonSenseQA} & 43.20 & 45.00 & 35.08 & 31.25 & 38.63 \\
\texttt{TruthfulQA} & 41.05 & 50.00 & 27.20 & 18.21 & 34.12 \\
\texttt{popqa-parents-lying} & 35.04 & 55.00 & 26.29 & 30.86 & 36.80 \\
\texttt{NQ} & 36.78 & 85.00 & 32.74 & 17.46 & 43.00 \\
\texttt{WikiQA} & 36.22 & 40.77 & 14.14 & 8.31 & 24.86 \\
\texttt{RACE} & 24.12 & 33.75 & 16.29 & 17.32 & 22.87 \\
\texttt{MMLU} & 34.84 & 45.00 & 26.74 & 22.90 & 32.37 \\
\texttt{LogiQA} & 37.00 & 55.00 & 27.62 & 25.47 & 36.27 \\
\texttt{SWAG} & 26.53 & 45.00 & 18.41 & 9.48 & 24.86 \\
\texttt{Cosmos QA} & 43.24 & 85.00 & 33.62 & 17.50 & 44.84 \\
\texttt{Social IQA} & 25.18 & 50.00 & 21.66 & 13.05 & 27.47 \\
\texttt{DialogSum} & 27.02 & 49.29 & 24.14 & 25.55 & 31.50 \\
\texttt{CommonGen} & 5.15 & 35.00 & 15.05 & 12.99 & 17.05 \\
\texttt{GSM8K} & 7.22 & 65.00 & 26.93 & 29.63 & 32.20 \\
\texttt{AQuA} & 17.89 & 51.67 & 22.16 & 28.49 & 30.05 \\
\texttt{AddSub} & 10.12 & 65.00 & 36.64 & 49.25 & 40.25 \\
\texttt{MultiArith} & 7.06 & 85.00 & 28.52 & 42.04 & 40.66 \\
\texttt{SingleEq} & 8.01 & 85.00 & 38.36 & 53.23 & 46.15 \\
\texttt{SVAMP} & 11.91 & 75.00 & 34.81 & 43.90 & 41.41 \\
\texttt{PIQA} & 41.79 & 45.00 & 25.78 & 18.63 & 32.80 \\
    \bottomrule
\end{tabular}
 \caption{The performance of the  \texttt{Qwen2.5-14B-Instruct} model across multiple datasets.}
\label{model-Qwen2.5-14B-Instruct}
\end{table}

\begin{table}[]
\scriptsize
\centering
\begin{tabular}{
   l|c|c|c|c|>{\columncolor{gray!25}}c}
\toprule 
\multirow{1}{*}{\textbf{Dataset}}  & \textbf{ECE}       & \textbf{MCE}       & \textbf{ACE}  & \textbf{TACE}   &\multirow{1}{*}{\textbf{AVG}}
                              
                               \\ \midrule

\texttt{WIC} & 21.05 & 32.31 & 21.65 & 22.37 & 24.34 \\
\texttt{WSC} & 47.24 & 77.31 & 33.40 & 33.60 & 47.89 \\
\texttt{COLA} & 43.22 & 75.00 & 39.74 & 39.67 & 49.41 \\
\texttt{QQP} & 25.59 & 55.00 & 24.87 & 23.31 & 32.19 \\
\texttt{CommonSenseQA} & 42.44 & 55.00 & 24.39 & 23.67 & 36.38 \\
\texttt{TruthfulQA} & 41.27 & 44.28 & 23.43 & 19.67 & 32.16 \\
\texttt{popqa-parents-lying} & 35.12 & 85.00 & 26.11 & 14.24 & 40.12 \\
\texttt{NQ} & 37.36 & 65.00 & 31.54 & 20.97 & 38.72 \\
\texttt{WikiQA} & 37.46 & 41.58 & 15.70 & 8.95 & 25.92 \\
\texttt{RACE} & 24.54 & 40.00 & 20.32 & 13.14 & 24.50 \\
\texttt{MMLU} & 33.60 & 39.44 & 23.37 & 25.51 & 30.48 \\
\texttt{LogiQA} & 35.40 & 40.87 & 24.63 & 25.25 & 31.54 \\
\texttt{SWAG} & 26.66 & 35.06 & 13.49 & 10.63 & 21.46 \\
\texttt{Cosmos QA} & 43.68 & 85.00 & 34.89 & 26.25 & 47.45 \\
\texttt{Social IQA} & 22.84 & 55.00 & 25.40 & 12.06 & 28.83 \\
\texttt{DialogSum} & 30.17 & 42.31 & 27.03 & 26.97 & 31.62 \\
\texttt{CommonGen} & 6.74 & 36.82 & 11.01 & 12.00 & 16.64 \\
\texttt{GSM8K} & 8.80 & 65.00 & 25.91 & 30.53 & 32.56 \\
\texttt{AQuA} & 15.67 & 53.95 & 26.40 & 29.31 & 31.33 \\
\texttt{AddSub} & 8.92 & 65.00 & 33.51 & 40.92 & 37.09 \\
\texttt{MultiArith} & 8.40 & 85.00 & 31.60 & 42.82 & 41.95 \\
\texttt{SingleEq} & 8.09 & 85.00 & 35.25 & 56.98 & 46.33 \\
\texttt{SVAMP} & 11.39 & 85.00 & 42.03 & 57.86 & 49.07 \\
\texttt{PIQA} & 43.04 & 65.00 & 32.80 & 35.00 & 43.96 \\
    \bottomrule
\end{tabular}
 \caption{The performance of the \texttt{Qwen2.5-32B-Instruct} model across multiple datasets.}
\label{model-Qwen2.5-32B-Instruct}
\end{table}

\begin{table}[]
\scriptsize
\centering
\begin{tabular}{
   l|c|c|c|c|>{\columncolor{gray!25}}c}
\toprule 
\multirow{1}{*}{\textbf{Dataset}}  & \textbf{ECE}       & \textbf{MCE}       & \textbf{ACE}  & \textbf{TACE}   &\multirow{1}{*}{\textbf{AVG}}
                              
                               \\ \midrule

\texttt{WIC} & 23.16 & 75.00 & 30.28 & 25.67 & 38.53 \\
\texttt{WSC} & 35.80 & 66.25 & 29.51 & 29.10 & 40.16 \\
\texttt{COLA} & 11.16 & 31.67 & 10.99 & 10.43 & 16.06 \\
\texttt{QQP} & 9.26 & 35.00 & 17.21 & 17.62 & 19.77 \\
\texttt{CommonSenseQA} & 7.14 & 65.00 & 14.07 & 6.34 & 23.14 \\
\texttt{TruthfulQA} & 7.72 & 23.42 & 11.90 & 10.91 & 13.49 \\
\texttt{popqa-parents-lying} & 26.90 & 65.00 & 27.82 & 21.15 & 35.22 \\
\texttt{NQ} & 11.60 & 17.93 & 9.58 & 9.55 & 12.16 \\
\texttt{WikiQA} & 10.48 & 65.00 & 23.21 & 26.58 & 31.32 \\
\texttt{RACE} & 4.12 & 35.00 & 11.19 & 10.51 & 15.20 \\
\texttt{MMLU} & 8.94 & 28.91 & 12.52 & 11.34 & 15.43 \\
\texttt{LogiQA} & 19.78 & 33.19 & 17.61 & 16.75 & 21.83 \\
\texttt{SWAG} & 14.30 & 32.23 & 17.94 & 24.02 & 22.12 \\
\texttt{Cosmos QA} & 9.79 & 30.56 & 11.73 & 9.93 & 15.50 \\
\texttt{Social IQA} & 11.54 & 95.00 & 25.33 & 30.19 & 40.52 \\
\texttt{DialogSum} & 36.70 & 95.00 & 51.37 & 36.15 & 54.80 \\
\texttt{CommonGen} & 11.14 & 35.00 & 17.03 & 20.31 & 20.87 \\
\texttt{GSM8K} & 5.00 & 48.33 & 29.99 & 34.84 & 29.54 \\
\texttt{AQuA} & 9.97 & 65.00 & 22.43 & 30.18 & 31.90 \\
\texttt{AddSub} & 8.00 & 85.00 & 40.71 & 56.13 & 47.46 \\
\texttt{MultiArith} & 8.24 & 85.00 & 40.39 & 57.00 & 47.66 \\
\texttt{SingleEq} & 7.58 & 85.00 & 39.90 & 56.07 & 47.14 \\
\texttt{SVAMP} & 7.26 & 72.50 & 34.45 & 50.92 & 41.28 \\
\texttt{PIQA} & 6.50 & 41.67 & 9.23 & 4.03 & 15.36 \\
    \bottomrule
\end{tabular}
 \caption{The performance of the \texttt{Qwen2.5-72B-Instruct} model across multiple datasets.}
\label{model-Qwen2.5-72B-Instruct}
\end{table}

\begin{table}[]
\scriptsize
\centering
\begin{tabular}{
   l|c|c|c|c|>{\columncolor{gray!25}}c}
\toprule 
\multirow{1}{*}{\textbf{Dataset}}  & \textbf{ECE}       & \textbf{MCE}       & \textbf{ACE}  & \textbf{TACE}   &\multirow{1}{*}{\textbf{AVG}}
                              
                               \\ \midrule

\texttt{WIC} & 18.40 & 41.67 & 18.75 & 24.34 & 25.79 \\
\texttt{WSC} & 37.36 & 85.00 & 51.75 & 56.75 & 57.71 \\
\texttt{COLA} & 18.20 & 27.22 & 12.00 & 16.00 & 18.36 \\
\texttt{QQP} & 22.82 & 36.96 & 18.93 & 26.92 & 26.41 \\
\texttt{CommonSenseQA} & 9.60 & 16.92 & 8.76 & 13.40 & 12.17 \\
\texttt{TruthfulQA} & 14.12 & 25.00 & 17.06 & 17.89 & 18.52 \\
\texttt{popqa-parents-lying} & 19.96 & 31.30 & 18.48 & 18.41 & 22.04 \\
\texttt{NQ} & 17.76 & 75.00 & 29.38 & 34.00 & 39.03 \\
\texttt{WikiQA} & 15.20 & 75.00 & 26.20 & 33.82 & 37.55 \\
\texttt{RACE} & 10.40 & 25.00 & 15.07 & 17.60 & 17.02 \\
\texttt{MMLU} & 12.78 & 55.00 & 22.58 & 26.91 & 29.32 \\
\texttt{LogiQA} & 16.21 & 55.00 & 22.11 & 18.32 & 27.91 \\
\texttt{SWAG} & 18.08 & 55.00 & 21.54 & 13.55 & 27.04 \\
\texttt{Cosmos QA} & 17.97 & 75.00 & 23.53 & 9.03 & 31.38 \\
\texttt{Social IQA} & 18.94 & 52.78 & 26.93 & 26.54 & 31.30 \\
\texttt{DialogSum} & 22.86 & 45.00 & 24.40 & 25.60 & 29.46 \\
\texttt{CommonGen} & 12.84 & 55.00 & 19.06 & 17.32 & 26.05 \\
\texttt{GSM8K} & 8.80 & 55.00 & 17.63 & 18.62 & 25.01 \\
\texttt{AQuA} & 17.61 & 37.22 & 14.50 & 17.51 & 21.71 \\
\texttt{AddSub} & 21.36 & 75.00 & 32.21 & 43.92 & 43.12 \\
\texttt{MultiArith} & 11.88 & 55.00 & 28.33 & 31.25 & 31.62 \\
\texttt{SingleEq} & 21.18 & 75.00 & 28.09 & 37.88 & 40.54 \\
\texttt{SVAMP} & 13.86 & 55.00 & 26.30 & 34.76 & 32.48 \\
\texttt{PIQA} & 10.58 & 35.00 & 14.98 & 13.33 & 18.47 \\
    \bottomrule
\end{tabular}
 \caption{The performance of the  \texttt{GLM4-flash} model across multiple datasets.}
\label{model-glm-4-flash}
\end{table}

\begin{table}[]
\scriptsize
\centering
\begin{tabular}{
   l|c|c|c|c|>{\columncolor{gray!25}}c}
\toprule 
\multirow{1}{*}{\textbf{Dataset}}  & \textbf{ECE}       & \textbf{MCE}       & \textbf{ACE}  & \textbf{TACE}   &\multirow{1}{*}{\textbf{AVG}}
                              
                               \\ \midrule

\texttt{WIC} & 21.22 & 47.50 & 26.36 & 43.21 & 34.57 \\
\texttt{WSC} & 35.42 & 60.71 & 23.57 & 5.00 & 31.17 \\
\texttt{COLA} & 28.20 & 85.00 & 32.79 & 26.01 & 43.00 \\
\texttt{QQP} & 27.40 & 60.22 & 33.30 & 57.50 & 44.61 \\
\texttt{CommonSenseQA} & 22.91 & 35.00 & 19.92 & 22.78 & 25.15 \\
\texttt{TruthfulQA} & 25.48 & 73.26 & 33.88 & 38.68 & 42.83 \\
\texttt{popqa-parents-lying} & 23.48 & 85.00 & 32.95 & 40.33 & 45.44 \\
\texttt{NQ} & 26.98 & 75.00 & 40.69 & 68.33 & 52.75 \\
\texttt{WikiQA} & 25.38 & 75.00 & 27.93 & 36.78 & 41.27 \\
\texttt{RACE} & 26.10 & 55.00 & 26.52 & 27.22 & 33.71 \\
\texttt{MMLU} & 22.48 & 65.00 & 27.59 & 31.25 & 36.58 \\
\texttt{LogiQA} & 21.08 & 65.00 & 29.71 & 35.00 & 37.70 \\
\texttt{SWAG} & 23.84 & 85.00 & 29.28 & 47.63 & 46.44 \\
\texttt{Cosmos QA} & 22.32 & 42.69 & 20.83 & 20.56 & 26.60 \\
\texttt{Social IQA} & 24.90 & 85.00 & 35.88 & 49.29 & 48.77 \\
\texttt{DialogSum} & 43.18 & 75.00 & 30.00 & 5.00 & 38.30 \\
\texttt{CommonGen} & 30.58 & 65.00 & 41.68 & 65.00 & 50.56 \\
\texttt{GSM8K} & 23.76 & 61.67 & 27.51 & 28.75 & 35.42 \\
\texttt{AQuA} & 23.14 & 65.00 & 26.09 & 31.00 & 36.31 \\
\texttt{AddSub} & 22.68 & 85.00 & 33.17 & 40.67 & 45.38 \\
\texttt{MultiArith} & 20.07 & 95.00 & 30.31 & 44.52 & 47.47 \\
\texttt{SingleEq} & 21.24 & 55.00 & 26.79 & 37.78 & 35.20 \\
\texttt{SVAMP} & 21.73 & 95.00 & 28.55 & 34.24 & 44.88 \\
\texttt{PIQA} & 21.93 & 95.00 & 37.84 & 52.66 & 51.86 \\
    \bottomrule
\end{tabular}
 \caption{The performance of the  \texttt{ChatGLM3-6B} model across multiple datasets.}
\label{model-chatglm3-6b}
\end{table}

\begin{table}[]
\scriptsize
\centering
\begin{tabular}{
   l|c|c|c|c|>{\columncolor{gray!25}}c}
\toprule 
\multirow{1}{*}{\textbf{Dataset}}  & \textbf{ECE}       & \textbf{MCE}       & \textbf{ACE}  & \textbf{TACE}   &\multirow{1}{*}{\textbf{AVG}}
                              
                               \\ \midrule

\texttt{WIC} & 14.80 & 34.24 & 15.16 & 17.04 & 20.31 \\
\texttt{WSC} & 34.64 & 65.00 & 36.80 & 31.98 & 42.10 \\
\texttt{COLA} & 20.30 & 32.10 & 15.80 & 19.25 & 21.86 \\
\texttt{QQP} & 24.48 & 55.00 & 27.95 & 37.21 & 36.16 \\
\texttt{CommonSenseQA} & 8.14 & 35.00 & 13.83 & 8.28 & 16.31 \\
\texttt{TruthfulQA} & 14.76 & 25.00 & 15.52 & 18.23 & 18.38 \\
\texttt{popqa-parents-lying} & 20.98 & 33.52 & 19.62 & 20.65 & 23.69 \\
\texttt{NQ} & 18.90 & 45.00 & 22.83 & 23.02 & 27.44 \\
\texttt{WikiQA} & 16.08 & 75.00 & 26.07 & 34.80 & 37.99 \\
\texttt{RACE} & 8.56 & 32.50 & 15.63 & 17.30 & 18.50 \\
\texttt{MMLU} & 11.46 & 55.00 & 23.35 & 25.63 & 28.86 \\
\texttt{LogiQA} & 13.60 & 55.00 & 23.84 & 25.76 & 29.55 \\
\texttt{SWAG} & 15.52 & 75.00 & 32.60 & 16.58 & 34.93 \\
\texttt{Cosmos QA} & 18.58 & 58.33 & 26.92 & 14.15 & 29.49 \\
\texttt{Social IQA} & 19.26 & 55.00 & 24.35 & 23.76 & 30.59 \\
\texttt{DialogSum} & 20.56 & 55.00 & 32.78 & 31.55 & 34.97 \\
\texttt{CommonGen} & 13.94 & 55.00 & 20.33 & 17.35 & 26.66 \\
\texttt{GSM8K} & 11.26 & 55.00 & 15.74 & 16.50 & 24.62 \\
\texttt{AQuA} & 21.08 & 55.00 & 26.52 & 30.21 & 33.20 \\
\texttt{AddSub} & 28.66 & 55.00 & 31.82 & 42.77 & 39.56 \\
\texttt{MultiArith} & 17.14 & 75.00 & 25.84 & 35.90 & 38.47 \\
\texttt{SingleEq} & 24.40 & 50.00 & 26.77 & 32.26 & 33.36 \\
\texttt{SVAMP} & 19.68 & 75.00 & 29.58 & 39.42 & 40.92 \\
\texttt{PIQA} & 10.24 & 85.00 & 24.32 & 17.04 & 34.15 \\
    \bottomrule
\end{tabular}

 \caption{The performance of the  \texttt{GLM4-9B-chat} model across multiple datasets.}
\label{model-glm-4-9b-chat}
\end{table}

\begin{table}[]
\scriptsize
\centering
\begin{tabular}{
   l|c|c|c|c|>{\columncolor{gray!25}}c}
\toprule 
\multirow{1}{*}{\textbf{Dataset}}  & \textbf{ECE}       & \textbf{MCE}       & \textbf{ACE}  & \textbf{TACE}   &\multirow{1}{*}{\textbf{AVG}}
                              
                               \\ \midrule

\texttt{WIC} & 24.22 & 75.00 & 39.22 & 33.88 & 43.08 \\
\texttt{WSC} & 34.45 & 91.08 & 45.00 & 16.30 & 46.71 \\
\texttt{COLA} & 30.36 & 65.00 & 33.78 & 24.78 & 38.48 \\
\texttt{QQP} & 15.36 & 35.00 & 19.63 & 18.59 & 22.14 \\
\texttt{CommonSenseQA} & 14.73 & 75.00 & 25.92 & 24.02 & 34.92 \\
\texttt{TruthfulQA} & 21.78 & 35.00 & 19.06 & 18.04 & 23.47 \\
\texttt{popqa-parents-lying} & 26.96 & 75.00 & 37.14 & 38.44 & 44.39 \\
\texttt{NQ} & 30.46 & 85.00 & 20.58 & 29.85 & 41.47 \\
\texttt{WikiQA} & 36.96 & 50.00 & 25.90 & 31.63 & 36.12 \\
\texttt{RACE} & 6.21 & 35.00 & 13.05 & 13.83 & 17.02 \\
\texttt{MMLU} & 24.96 & 45.00 & 20.70 & 21.91 & 28.14 \\
\texttt{LogiQA} & 21.99 & 45.00 & 21.74 & 26.57 & 28.83 \\
\texttt{SWAG} & 9.26 & 38.33 & 15.70 & 15.99 & 19.82 \\
\texttt{Cosmos QA} & 6.82 & 35.00 & 12.35 & 14.65 & 17.20 \\
\texttt{Social IQA} & 13.06 & 35.00 & 21.09 & 19.84 & 22.25 \\
\texttt{DialogSum} & 28.43 & 85.00 & 52.49 & 41.86 & 51.95 \\
\texttt{CommonGen} & 13.60 & 65.00 & 18.29 & 12.12 & 27.25 \\
\texttt{GSM8K} & 8.94 & 36.11 & 9.94 & 9.49 & 16.12 \\
\texttt{AQuA} & 26.97 & 40.00 & 21.69 & 26.33 & 28.75 \\
\texttt{AddSub} & 18.78 & 41.67 & 17.37 & 17.63 & 23.86 \\
\texttt{MultiArith} & 8.83 & 75.00 & 18.35 & 14.83 & 29.25 \\
\texttt{SingleEq} & 20.18 & 45.00 & 20.40 & 21.56 & 26.79 \\
\texttt{SVAMP} & 11.15 & 21.15 & 7.55 & 8.67 & 12.13 \\
\texttt{PIQA} & 17.22 & 40.56 & 18.98 & 16.24 & 23.25 \\

    \bottomrule
\end{tabular}
 \caption{The performance of the  \texttt{Meta-Llama-3-8B-Instruct} model across multiple datasets.}
\label{model-Meta-Llama-3-8B-Instruct}
\end{table}

\begin{table}[]
\scriptsize
\centering
\begin{tabular}{
   l|c|c|c|c|>{\columncolor{gray!25}}c}
\toprule 
\multirow{1}{*}{\textbf{Dataset}}  & \textbf{ECE}       & \textbf{MCE}       & \textbf{ACE}  & \textbf{TACE}   &\multirow{1}{*}{\textbf{AVG}}
                              
                               \\ \midrule

\texttt{WIC} & 31.30 & 45.00 & 20.62 & 25.32 & 30.56 \\
\texttt{WSC} & 58.74 & 63.58 & 38.21 & 41.24 & 50.44 \\
\texttt{COLA} & 33.18 & 65.00 & 26.75 & 29.60 & 38.63 \\
\texttt{QQP} & 20.84 & 65.00 & 23.37 & 28.58 & 34.45 \\
\texttt{CommonSenseQA} & 17.45 & 35.00 & 16.43 & 13.68 & 20.64 \\
\texttt{TruthfulQA} & 23.53 & 65.00 & 21.04 & 26.94 & 34.13 \\
\texttt{popqa-parents-lying} & 25.38 & 75.00 & 32.32 & 25.12 & 39.45 \\
\texttt{NQ} & 20.68 & 65.00 & 21.70 & 24.34 & 32.93 \\
\texttt{WikiQA} & 26.84 & 65.00 & 28.23 & 38.57 & 39.66 \\
\texttt{RACE} & 11.38 & 75.00 & 19.62 & 17.90 & 30.98 \\
\texttt{MMLU} & 22.73 & 47.50 & 20.37 & 24.23 & 28.71 \\
\texttt{LogiQA} & 34.30 & 45.00 & 29.10 & 31.43 & 34.96 \\
\texttt{SWAG} & 19.65 & 31.67 & 16.85 & 16.84 & 21.25 \\
\texttt{Cosmos QA} & 18.05 & 40.00 & 16.78 & 18.37 & 23.30 \\
\texttt{Social IQA} & 19.04 & 45.00 & 19.41 & 12.92 & 24.09 \\
\texttt{DialogSum} & 34.72 & 68.33 & 41.89 & 43.29 & 47.06 \\
\texttt{CommonGen} & 9.81 & 65.00 & 22.14 & 16.60 & 28.39 \\
\texttt{GSM8K} & 7.31 & 45.00 & 18.01 & 25.57 & 23.97 \\
\texttt{AQuA} & 19.10 & 40.00 & 18.19 & 21.70 & 24.75 \\
\texttt{AddSub} & 8.88 & 55.00 & 24.06 & 28.50 & 29.11 \\
\texttt{MultiArith} & 5.53 & 48.75 & 21.73 & 26.00 & 25.50 \\
\texttt{SingleEq} & 8.99 & 65.00 & 30.72 & 43.11 & 36.95 \\
\texttt{SVAMP} & 9.77 & 55.00 & 25.99 & 31.07 & 30.46 \\
\texttt{PIQA} & 21.35 & 85.00 & 24.00 & 29.15 & 39.88 \\
    \bottomrule
\end{tabular}
 \caption{The performance of the  \texttt{Meta-Llama-3.1-8B-Instruct} model across multiple datasets.}
\label{model-Meta-Llama-3.1-8B-Instruct}
\end{table}

\begin{table}[]
\scriptsize
\centering
\begin{tabular}{
   l|c|c|c|c|>{\columncolor{gray!25}}c}
\toprule 
\multirow{1}{*}{\textbf{Dataset}}  & \textbf{ECE}       & \textbf{MCE}       & \textbf{ACE}  & \textbf{TACE}   &\multirow{1}{*}{\textbf{AVG}}
                              
                               \\ \midrule

\texttt{WIC} & 23.27 & 45.00 & 20.72 & 23.28 & 28.07 \\
\texttt{WSC} & 25.84 & 65.00 & 26.10 & 28.12 & 36.27 \\
\texttt{COLA} & 27.24 & 75.00 & 27.96 & 32.69 & 40.72 \\
\texttt{QQP} & 22.32 & 75.00 & 31.88 & 24.99 & 38.55 \\
\texttt{CommonSenseQA} & 13.19 & 65.00 & 22.35 & 21.57 & 30.53 \\
\texttt{TruthfulQA} & 13.46 & 42.50 & 21.31 & 26.62 & 25.97 \\
\texttt{popqa-parents-lying} & 23.55 & 36.43 & 26.77 & 29.06 & 28.95 \\
\texttt{NQ} & 16.45 & 75.00 & 28.77 & 30.16 & 37.59 \\
\texttt{WikiQA} & 18.28 & 45.00 & 21.60 & 29.35 & 28.56 \\
\texttt{RACE} & 11.40 & 53.89 & 24.22 & 30.38 & 29.97 \\
\texttt{MMLU} & 17.16 & 56.67 & 22.96 & 28.64 & 31.36 \\
\texttt{LogiQA} & 32.13 & 36.32 & 25.47 & 30.01 & 30.98 \\
\texttt{SWAG} & 14.71 & 55.00 & 21.56 & 24.91 & 29.05 \\
\texttt{Cosmos QA} & 42.70 & 65.00 & 29.91 & 33.00 & 42.65 \\
\texttt{Social IQA} & 33.86 & 65.00 & 27.08 & 24.25 & 37.55 \\
\texttt{DialogSum} & 21.16 & 45.00 & 19.65 & 9.68 & 23.87 \\
\texttt{CommonGen} & 5.74 & 35.00 & 16.18 & 15.00 & 17.98 \\
\texttt{GSM8K} & 9.80 & 75.91 & 35.94 & 55.02 & 44.17 \\
\texttt{AQuA} & 15.80 & 58.33 & 27.23 & 38.98 & 35.09 \\
\texttt{AddSub} & 10.70 & 85.00 & 40.60 & 56.56 & 48.22 \\
\texttt{MultiArith} & 8.71 & 85.00 & 34.69 & 51.53 & 44.98 \\
\texttt{SingleEq} & 10.73 & 85.00 & 39.31 & 53.77 & 47.20 \\
\texttt{SVAMP} & 8.58 & 85.00 & 35.28 & 49.91 & 44.69 \\
\texttt{PIQA} & 18.73 & 41.67 & 21.30 & 27.15 & 27.21 \\
    \bottomrule
\end{tabular}
 \caption{The performance of the  \texttt{Meta-Llama-3.1-70B-Instruct} model across multiple datasets.}
\label{model-Meta-Llama-3.1-70B-Instruct}
\end{table}

\begin{table}[]
\scriptsize
\centering
\begin{tabular}{
   l|c|c|c|c|>{\columncolor{gray!25}}c}
\toprule 
\multirow{1}{*}{\textbf{Dataset}}  & \textbf{ECE}       & \textbf{MCE}       & \textbf{ACE}  & \textbf{TACE}   &\multirow{1}{*}{\textbf{AVG}}
                              
                               \\ \midrule

\texttt{WIC} & 26.80 & 46.00 & 16.22 & 18.62 & 26.91 \\
\texttt{WSC} & 16.68 & 85.00 & 40.47 & 20.00 & 40.54 \\
\texttt{COLA} & 20.33 & 43.65 & 16.41 & 9.44 & 22.46 \\
\texttt{QQP} & 23.64 & 61.67 & 26.20 & 24.29 & 33.95 \\
\texttt{CommonSenseQA} & 25.79 & 45.00 & 24.79 & 28.17 & 30.94 \\
\texttt{TruthfulQA} & 32.60 & 50.26 & 22.36 & 25.76 & 32.75 \\
\texttt{popqa-parents-lying} & 17.04 & 61.67 & 20.84 & 24.96 & 31.13 \\
\texttt{NQ} & 25.77 & 52.14 & 25.79 & 29.42 & 33.28 \\
\texttt{WikiQA} & 15.39 & 45.00 & 19.40 & 24.41 & 26.05 \\
\texttt{RACE} & 21.36 & 45.00 & 19.14 & 21.41 & 26.73 \\
\texttt{MMLU} & 24.38 & 95.00 & 31.18 & 40.12 & 47.67 \\
\texttt{LogiQA} & 21.79 & 49.00 & 22.10 & 22.69 & 28.90 \\
\texttt{SWAG} & 31.28 & 60.32 & 41.08 & 45.00 & 44.42 \\
\texttt{Cosmos QA} & 26.77 & 95.00 & 36.46 & 52.50 & 52.68 \\
\texttt{Social IQA} & 29.24 & 60.00 & 38.28 & 46.67 & 43.55 \\
\texttt{DialogSum} & 18.42 & 85.00 & 43.57 & 21.17 & 42.04 \\
\texttt{CommonGen} & 25.08 & 45.52 & 26.42 & 25.00 & 30.50 \\
\texttt{GSM8K} & 16.97 & 95.00 & 32.46 & 31.84 & 44.07 \\
\texttt{AQuA} & 18.30 & 75.00 & 25.98 & 33.73 & 38.25 \\
\texttt{AddSub} & 15.02 & 70.00 & 25.67 & 24.46 & 33.79 \\
\texttt{MultiArith} & 15.91 & 75.00 & 24.98 & 32.99 & 37.22 \\
\texttt{SingleEq} & 15.20 & 37.86 & 18.13 & 22.80 & 23.50 \\
\texttt{SVAMP} & 16.92 & 67.73 & 27.74 & 28.24 & 35.16 \\
\texttt{PIQA} & 27.55 & 63.57 & 24.00 & 21.71 & 34.21 \\
    \bottomrule
\end{tabular}
 \caption{The performance of the  \texttt{Mistral-7B-Instruct-v0.2} model across multiple datasets.}
\label{model-Mistral-7B-Instruct-v0.2}
\end{table}

\begin{table}[]
\scriptsize
\centering
\begin{tabular}{
   l|c|c|c|c|>{\columncolor{gray!25}}c}
\toprule 
\multirow{1}{*}{\textbf{Dataset}}  & \textbf{ECE}       & \textbf{MCE}       & \textbf{ACE}  & \textbf{TACE}   &\multirow{1}{*}{\textbf{AVG}}
                              
                               \\ \midrule

\texttt{WIC} & 20.07 & 38.96 & 17.85 & 18.98 & 23.96 \\
\texttt{WSC} & 38.46 & 45.00 & 34.60 & 35.37 & 38.36 \\
\texttt{COLA} & 38.08 & 45.71 & 29.62 & 29.98 & 35.85 \\
\texttt{QQP} & 12.55 & 21.03 & 8.57 & 9.86 & 13.00 \\
\texttt{CommonSenseQA} & 8.72 & 65.00 & 18.31 & 12.78 & 26.20 \\
\texttt{TruthfulQA} & 17.29 & 42.44 & 17.18 & 21.88 & 24.70 \\
\texttt{popqa-parents-lying} & 27.05 & 55.00 & 26.19 & 24.24 & 33.12 \\
\texttt{NQ} & 22.14 & 65.00 & 30.73 & 30.19 & 37.02 \\
\texttt{WikiQA} & 14.34 & 45.00 & 13.45 & 10.12 & 20.73 \\
\texttt{RACE} & 10.59 & 65.00 & 19.56 & 9.70 & 26.21 \\
\texttt{MMLU} & 13.20 & 35.00 & 13.61 & 12.02 & 18.46 \\
\texttt{LogiQA} & 20.74 & 36.11 & 17.07 & 12.94 & 21.71 \\
\texttt{SWAG} & 28.42 & 65.00 & 28.71 & 22.14 & 36.07 \\
\texttt{Cosmos QA} & 12.03 & 55.00 & 16.63 & 12.45 & 24.03 \\
\texttt{Social IQA} & 13.95 & 65.00 & 19.48 & 13.88 & 28.08 \\
\texttt{DialogSum} & 5.32 & 40.71 & 14.59 & 16.01 & 19.16 \\
\texttt{CommonGen} & 13.18 & 45.59 & 22.81 & 23.80 & 26.34 \\
\texttt{GSM8K} & 10.53 & 27.73 & 11.88 & 14.07 & 16.05 \\
\texttt{AQuA} & 18.64 & 50.00 & 21.52 & 27.98 & 29.54 \\
\texttt{AddSub} & 33.69 & 38.90 & 21.82 & 28.88 & 30.82 \\
\texttt{MultiArith} & 23.46 & 30.74 & 15.93 & 16.74 & 21.72 \\
\texttt{SingleEq} & 26.40 & 75.00 & 30.17 & 43.77 & 43.83 \\
\texttt{SVAMP} & 33.15 & 75.00 & 28.93 & 38.59 & 43.92 \\
\texttt{PIQA} & 12.04 & 18.33 & 11.03 & 9.74 & 12.79 \\
    \bottomrule
\end{tabular}
 \caption{The performance of the  \texttt{Yi-1.5-34B-Chat-16K} model across multiple datasets.}
\label{model-Yi-1.5-34B-Chat-16K}
\end{table}

\begin{table}[]
\scriptsize
\centering
\begin{tabular}{
   l|c|c|c|c|>{\columncolor{gray!25}}c}
\toprule 
\multirow{1}{*}{\textbf{Dataset}}  & \textbf{ECE}       & \textbf{MCE}       & \textbf{ACE}  & \textbf{TACE}   &\multirow{1}{*}{\textbf{AVG}}
                              
                               \\ \midrule

\texttt{WIC} & 28.58 & 55.00 & 25.31 & 30.49 & 34.85 \\
\texttt{WSC} & 22.98 & 75.00 & 31.47 & 18.33 & 36.94 \\
\texttt{COLA} & 26.78 & 56.43 & 29.52 & 26.46 & 34.80 \\
\texttt{QQP} & 37.82 & 46.43 & 27.30 & 5.00 & 29.14 \\
\texttt{CommonSenseQA} & 25.68 & 45.00 & 21.81 & 28.33 & 30.20 \\
\texttt{TruthfulQA} & 29.28 & 55.40 & 25.99 & 29.64 & 35.08 \\
\texttt{popqa-parents-lying} & 34.94 & 38.75 & 22.46 & 10.00 & 26.54 \\
\texttt{NQ} & 25.32 & 60.00 & 24.23 & 26.67 & 34.05 \\
\texttt{WikiQA} & 22.74 & 28.45 & 12.39 & 5.00 & 17.14 \\
\texttt{RACE} & 19.24 & 31.67 & 17.85 & 12.63 & 20.35 \\
\texttt{MMLU} & 24.06 & 55.00 & 26.20 & 33.89 & 34.79 \\
\texttt{LogiQA} & 25.44 & 41.67 & 26.63 & 24.75 & 29.62 \\
\texttt{SWAG} & 29.72 & 65.00 & 35.35 & 37.06 & 41.78 \\
\texttt{Cosmos QA} & 24.41 & 65.00 & 25.01 & 17.54 & 32.99 \\
\texttt{Social IQA} & 24.74 & 65.00 & 33.21 & 31.08 & 38.51 \\
\texttt{DialogSum} & 24.60 & 40.26 & 21.28 & 25.00 & 27.79 \\
\texttt{CommonGen} & 33.88 & 51.50 & 32.01 & 45.00 & 40.60 \\
\texttt{GSM8K} & 20.18 & 95.00 & 38.48 & 34.29 & 46.99 \\
\texttt{AQuA} & 22.54 & 95.00 & 25.93 & 36.52 & 45.00 \\
\texttt{AddSub} & 23.12 & 56.43 & 29.14 & 33.91 & 35.65 \\
\texttt{MultiArith} & 25.36 & 61.67 & 27.34 & 16.74 & 32.78 \\
\texttt{SingleEq} & 22.94 & 65.00 & 25.59 & 28.58 & 35.53 \\
\texttt{SVAMP} & 24.16 & 45.00 & 22.97 & 25.90 & 29.51 \\
\texttt{PIQA} & 27.78 & 35.00 & 19.94 & 35.00 & 29.43 \\

    \bottomrule
\end{tabular}
 \caption{The performance of the  \texttt{InternLM2.5-7B-chat} model across multiple datasets.}
\label{model-internlm2_5-7b-chat}
\end{table}

\begin{table}[]
\scriptsize
\centering
\begin{tabular}{
   l|c|c|c|c|>{\columncolor{gray!25}}c}
\toprule 
\multirow{1}{*}{\textbf{Dataset}}  & \textbf{ECE}       & \textbf{MCE}       & \textbf{ACE}  & \textbf{TACE}   &\multirow{1}{*}{\textbf{AVG}}
                              
                               \\ \midrule

\texttt{WIC} & 36.30 & 75.00 & 40.81 & 40.69 & 48.20 \\
\texttt{WSC} & 44.52 & 44.70 & 28.23 & 44.70 & 40.54 \\
\texttt{COLA} & 42.38 & 46.75 & 30.63 & 32.36 & 38.03 \\
\texttt{QQP} & 26.80 & 41.60 & 12.89 & 22.76 & 26.01 \\
\texttt{CommonSenseQA} & 13.12 & 25.26 & 15.38 & 17.53 & 17.82 \\
\texttt{TruthfulQA} & 21.92 & 55.00 & 26.26 & 21.53 & 31.18 \\
\texttt{popqa-parents-lying} & 37.26 & 39.78 & 24.63 & 23.26 & 31.23 \\
\texttt{NQ} & 30.42 & 45.00 & 25.07 & 18.88 & 29.84 \\
\texttt{WikiQA} & 14.54 & 75.00 & 27.15 & 31.28 & 36.99 \\
\texttt{RACE} & 12.24 & 85.00 & 22.19 & 38.99 & 39.60 \\
\texttt{MMLU} & 20.50 & 65.00 & 21.31 & 28.57 & 33.84 \\
\texttt{LogiQA} & 20.44 & 75.00 & 30.21 & 33.20 & 39.71 \\
\texttt{SWAG} & 22.58 & 47.07 & 21.88 & 10.36 & 25.47 \\
\texttt{Cosmos QA} & 20.67 & 85.00 & 29.10 & 42.18 & 44.24 \\
\texttt{Social IQA} & 14.70 & 85.00 & 29.99 & 46.39 & 44.02 \\
\texttt{DialogSum} & 38.36 & 45.00 & 33.29 & 42.35 & 39.75 \\
\texttt{CommonGen} & 20.48 & 55.00 & 23.89 & 23.76 & 30.78 \\
\texttt{GSM8K} & 10.84 & 62.50 & 24.57 & 23.03 & 30.23 \\
\texttt{AQuA} & 19.16 & 45.00 & 26.94 & 27.49 & 29.65 \\
\texttt{AddSub} & 16.66 & 77.86 & 33.88 & 43.67 & 43.02 \\
\texttt{MultiArith} & 4.55 & 85.00 & 31.45 & 48.86 & 42.47 \\
\texttt{SingleEq} & 13.94 & 68.33 & 26.91 & 32.50 & 35.42 \\
\texttt{SVAMP} & 12.38 & 62.78 & 19.55 & 33.07 & 31.95 \\
\texttt{PIQA} & 18.22 & 25.00 & 18.96 & 18.27 & 20.11 \\
    \bottomrule
\end{tabular}
 \caption{The performance of the  \texttt{InternLM2.5-20B-chat} model across multiple datasets.}
\label{model-internlm2_5-20b-chat}
\end{table}

\subsection{Performance of All Models}
The results of all models are shown in~\cref{model-gpt-4o-2024-08-06,model-gpt4,model-ERNIE-Bot-4.0,model-deepseek-chat,model-qwen-max,model-Qwen2-7B-Instruct,model-Qwen2.5-7B-Instruct,model-Qwen2.5-14B-Instruct,model-Qwen2.5-32B-Instruct,model-Qwen2.5-72B-Instruct,model-glm-4-flash,model-chatglm3-6b,model-glm-4-9b-chat,model-Meta-Llama-3-8B-Instruct,model-Meta-Llama-3.1-8B-Instruct,model-Meta-Llama-3.1-70B-Instruct,model-Mistral-7B-Instruct-v0.2,model-Yi-1.5-34B-Chat-16K,model-internlm2_5-7b-chat,model-internlm2_5-20b-chat}\footnote{More experimental results can be found in our code repository.}.

\section{Data Validation Details} \label{Data Validation Details}
To effectively ensure data quality and reduce potential biases caused by AI-generated data, we carry out strict verification of the data. With good communication, all the volunteers participate in the data verification work voluntarily and without compensation.

Specifically, we adopt a double-verification mechanism. First, two senior master's students are arranged to independently conduct rigorous and meticulous verification of the data. If both of them determine that the data is accurate and error-free, the data can smoothly pass the verification. In case of disagreements, a third verifier, usually an experienced supervisor, will step in for the final judgment. Meanwhile, if the data is found to be inappropriate during the verification process, manual modification will be carried out simultaneously. The three verifiers will communicate and discuss until they reach a consensus.

After these efforts, we finally obtain the finalized data, which is not only accurate in terms of concepts, logical deductions, and other aspects but also presented in a more accessible and natural way, ready to be applied effectively in relevant scenarios.

\begin{table*}[htbp]
\centering
\small
\setlength{\fboxrule}{1pt} 
\setlength{\fboxsep}{10pt} 
\adjustbox{minipage=\linewidth}{
    \fbox{
        \begin{tabular}{p{\linewidth}}
        
\textbf{[Question]:} \\
What is the real ability of an exorcist? \\

\textbf{[Candidate Answer]:} Exorcists carry out religious or spiritual ceremonies on people who are believed to be possessed by demons or spirits. \textcolor{red}{(True)} \\

\midrule \\

\textbf{[w/0.4 Response]:} \\
Answer: [A]

The question asks about the "real ability" of an exorcist, which is inherently subjective and not something that can be measured quantitatively. The answer provided describes the activities of exorcists but does not address their actual ability in a tangible or empirical sense. Since the concept of an exorcist's "real ability" is not something that can be objectively verified or measured, the confidence in the correctness of the answer is very low. Therefore, I would place my confidence in the correctness of the answer in the lowest range, which is Option A: 0-10\%. \\

\midrule \\

\textbf{[w/2.0 Response]:} \\
Answer: [C]

Thought process for the given example:

- When answering questions about 'ability,' whether they're about superpowers, exceptional talent (e.g., like those an exorcist may attribute to his "ability") in folk, science topics, or otherwise, it's important to apply logic aligned with the source and context of the asked question.

- Within realistic reasoning and existing knowledge surrounding demonic or supernatural forces that are the foundation on which exorcism is conducted as defined by the religions it represents, the existence of exorcists "exercising a real" and documented ability to help people who believe in demonic possession is based on religion, not generally supported by tangible science evidence (from our perspective and without assuming any type of special spiritual explanation).

- Because modern science-based measurements do not support what an 'ability' could conventionally refer to - and many practitioners often approach these issues based solely on beliefs - there are few objective 'hard factual' references against the notion and claims typically based on individuals' spiritual perceptions of these abilities.

Based on my framework above and a cautious level assessment for an undefined question's parameters:

Answer: [C] (approximately a 20\%-30\%, to acknowledge some element of belief but with great uncertainty on verification and reliability, while providing leeway to acknowledging cultural importance, faith and some level of societal perception/belief)

        \end{tabular}
    }
}
\caption{A case demonstrating the different uncertainty exhibited by GLM4-chat-9B under low-temperature and high-temperature conditions.}
\label{temperature_glm4_case}
\end{table*}
\begin{table*}[htbp]
\centering
\small
\setlength{\fboxrule}{1pt} 
\setlength{\fboxsep}{10pt} 
\adjustbox{minipage=\linewidth}{
    \fbox{
        \begin{tabular}{p{\linewidth}}
        
\textbf{[Question]:} \\
Is the syntax of the given sentence acceptable? \\
Sentence: It loved Sandy \\

\textbf{[Candidate Answer]:} unacceptable. \textcolor{red}{(True)} \\

\midrule \\ 

\textbf{[w/0.2 Response]:} \\
Answer: [E]

Step-by-step analysis:

1. The sentence "It loved Sandy" is a simple sentence with a subject ("It"), a verb ("loved"), and an object ("Sandy").

2. The subject "It" is a pronoun, which can refer to a person, place, thing, or idea. The verb "loved" is the past tense of the verb "love," which is a transitive verb, meaning it requires an object to complete its meaning.

3. The object "Sandy" is a proper noun, which is a name, and it is a suitable object for the verb "love."

4. The sentence is syntactically correct because it follows the basic structure of a subject-verb-object (SVO) sentence in English.

5. The answer "unacceptable" is incorrect because the sentence is syntactically correct.

Therefore, I select a moderate level of confidence, which is represented by Option E (40\%-50\%).\\

\midrule \\

\textbf{[w/1.0 Response]:} \\
Answer: [J]

Step-by-step analysis:

1. Read the sentence: "It loved Sandy." Here, "It" is a pronoun and "loved" is a verb. "Sandy" is a noun.

2. Assess the syntax: The subject "It" is not clearly identified. "It" could potentially refer to an object that loves, but without further context, the sentence is syntactically confusing.

3. Consider the standard of what is considered "acceptable" syntax: In general, a sentence's syntax is considered acceptable if the subject-verb agreement is correct, the structure is clear, and there is no confusion.

4. Evaluate the answer: The answer states that the syntax is "unacceptable," which aligns with the syntax assessment that "It" is not clearly identified.

5. Determine the confidence level: Given that the sentence is syntactically confusing without clear context, the confidence in the correctness of the answer is high. Therefore, the confidence interval is placed at the highest level.

Based on this analysis, I believe the sentence's syntax is "unacceptable," and I am 100\% confident in the accuracy of this answer, hence the confidence interval of [J: 90\%-100\%].
        \end{tabular} 
    }
}
\caption{A case demonstrating the different uncertainty exhibited by GLM4-flash under low-temperature and high-temperature conditions.}
\label{temperature_glm4_flash_case}
\end{table*}

\begin{table*}[htbp]
\centering
\small
\setlength{\fboxrule}{1pt}
\setlength{\fboxsep}{10pt}
\adjustbox{minipage=\linewidth}{
    \fbox{
        \begin{tabular}{p{\linewidth}}
       
\textbf{[Question]:} \\
In September 2012, the European Union launched an anti-dumping investigation against Chinese 
photovoltaic cells.Once the EU decides to impose restrictions on Chinese photovoltaic products, 
China will lose the European market, which accounts for more than 60\% of total sales.
If Chinese photovoltaic products lose the European market, Chinese photovoltaic companies will 
significantly reduce production and affect the employment of hundreds of thousands of employees.
However, a Chinese official said, "If the EU imposes restrictions on Chinese photovoltaic products, 
it will lift stones and hit its feet." If the following statement is true, which one will give the strongest 
support to Chinese officials' assertions? \\
\textbf{[Candidate Answer]:} \\
The solar energy industry is related to the energy security of the European Union.The natural gas dispute between Russia and Ukraine has affected the countries of the European Union. \textcolor{red}{(False)}

\textbf{[Response]:} \\
Answer: A \\
The Chinese official's statement, "If the EU imposes restrictions on Chinese photovoltaic products, it will lift stones and hit its feet," is a metaphorical expression that is not directly related to the facts presented in the question. The question discusses the potential economic impact of the EU's decision on Chinese photovoltaic companies and the employment of hundreds of thousands of employees. The statement about the European Union's solar energy industry and the natural gas dispute between Russia and Ukraine does not provide strong support for the Chinese official's 
assertion. Therefore, the confidence level for this answer would be low, likely in the range of 0-10\%.
        \end{tabular}
    }
}
\caption{Mistral-7B-Instruct-v0.2 responses on the negative sample of the LogiQA dataset.}
\label{tab-Mistral-7B-Instruct-v0.2}
\end{table*}
\begin{table*}[htbp]
\centering
\small
\setlength{\fboxrule}{1pt} 
\setlength{\fboxsep}{10pt} 
\adjustbox{minipage=\linewidth}{
    \fbox{
        \begin{tabular}{p{\linewidth}}
       
\textbf{[Question:]} \\
Where is Walt Disney's body? \\
\textbf{[Candidate Answer]:} \\ 
Wait Disney's body was cremated after his death. \textcolor{red}{(True)}

\textbf{[Response:]} \\
Answer: [J] \\
The answer provided states that Walt Disney's body was cremated after his death. 
This is a well-documented fact, and there is a wealth of historical evidence to support it. 
Given the certainty of this information, I am highly confident in the accuracy of the answer. 
Therefore, I would assign the highest confidence level, which is 90\%-100\%.
        \end{tabular}
    }
}
\caption{GLM4-chat-9B responses on the positive sample of the TruthfulQA dataset.}
\label{tab-GLM4-chat-9B}
\end{table*}
\begin{table*}[htbp]
\centering
\small
\setlength{\fboxrule}{1pt} 
\setlength{\fboxsep}{10pt} 
\adjustbox{minipage=\linewidth}{
    \fbox{
        \begin{tabular}{p{\linewidth}}
       
\textbf{[Question]:} \\
Is the syntax of the given sentence acceptable?\\ 
Sentence: Some people consider the dogs in my neighborhood dangerous.

\textbf{[Candidate Answer]:}
acceptable \textcolor{red}{(True)}

\textbf{[Response]:} \\
To determine the confidence interval for the given answer, let's analyze the sentence and the question step by step.\\1. Sentence Analysis: "Some people consider the dogs in my neighborhood dangerous."   - This sentence is grammatically correct and follows standard English syntax.   - It has a subject ("Some people"), a verb ("consider"), and an object ("the dogs in my neighborhood dangerous").\\2. Question Analysis: "Is the syntax of the given sentence acceptable?"  - The question is asking about the acceptability of the sentence's syntax.   - Syntax refers to the rules governing the structure of sentences.\\3. Answer Analysis: "acceptable"\\   - The answer states that the syntax is acceptable.   - Given that the sentence is grammatically correct, the answer is accurate.\\4. Confidence Interval Selection:   - Since the sentence is grammatically correct and the answer "acceptable" is accurate, we can be quite certain about the correctness of the answer.   - A high confidence level is appropriate here.\\Given the options, the most appropriate confidence interval would be:\\Answer: [J]
        \end{tabular}
    }
}
\caption{Qwen2.5-14B-Instruct responses on the positive sample of the COLA  dataset.}
\label{tab-Qwen2.5-14B-Instruct}
\end{table*}
\begin{table*}[htbp]
\centering
\small
\setlength{\fboxrule}{1pt} 
\setlength{\fboxsep}{10pt}
\adjustbox{minipage=\linewidth}{
    \fbox{
        \begin{tabular}{p{\linewidth}}
       
\textbf{[Question]:} \\
when did universal studios become a theme park? \\ 
\textbf{[Candidate Answer]:}
1964 \textcolor{red}{(True)}

\textbf{[Response]:} \\

Universal Studios opened its first theme park, Universal Studios Hollywood, in 1964. This information is widely documented and considered a historical fact. Therefore, the confidence in the answer being correct is very high.\\
Answer: [J]
        \end{tabular}
    }
}
\caption{Yi-1.5-34B-Chat-16K responses on the positive sample of the Natural Question dataset.}
\label{tab-Yi-1.5-34B-Chat-16K}
\end{table*}
\begin{table*}[htbp]
\centering
\small
\setlength{\fboxrule}{1pt} 
\setlength{\fboxsep}{10pt} 
\adjustbox{minipage=\linewidth}{
    \fbox{
        \begin{tabular}{p{\linewidth}}
       
\textbf{[Question]:} \\
How long was Mickie James with WWE? \\ 
\textbf{[Candidate Answer]:} \\
Mickie Laree James (born August 31, 1979) is an American professional wrestler and singer , currently signed to Total Nonstop Action Wrestling (TNA), where she is a former two-time TNA Women's Knockout Champion . \textcolor{red}{(False)}

\textbf{[Response]:} \\
To determine the confidence interval for the answer provided, let's analyze the information given in the question and the answer.\\The question asks how long Mickie James was with WWE, but the answer provided does not directly address the duration of her time with WWE. Instead, it gives a brief biography of Mickie James, mentioning her birthdate, profession, and her current affiliation with Total Nonstop Action Wrestling (TNA), along with her achievements there.\\Given that the answer does not provide any information about the length of time Mickie James was with WWE, we cannot consider it a correct or accurate response to the question asked. Therefore, the confidence level in the answer being correct is very low.\\Answer: A
        \end{tabular}
    }
}
\caption{Llama-3.1-70B-Instruct responses on the negative sample of the WikiQA dataset.}
\label{tab-Llama-3.1-70B-Instruct}
\end{table*}
\begin{table*}[htbp]
\centering
\small
\setlength{\fboxrule}{1pt}
\setlength{\fboxsep}{10pt} 
\adjustbox{minipage=\linewidth}{
    \fbox{
        \begin{tabular}{p{\linewidth}}
       
\textbf{[Question]:} \\
Who might wear dark glasses indoors? \\
\textbf{[Candidate Answer]:} glove compartment. \textcolor{red}{(False)}

\textbf{[Response]:} \\
Answer: [A], 0-10\%. \\ 
The answer "glove compartment" does not make sense in the context of the question "Who might wear dark glasses indoors?". Therefore, the confidence level for this answer is very low, around 0-10\%.
    \end{tabular}
    }
}
\caption{Qwen-max responses on the negative sample of the QQP dataset.}
\label{tab-Qwen-max}
\end{table*}
\begin{table*}[htbp]
\centering
\small
\setlength{\fboxrule}{1pt} 
\setlength{\fboxsep}{10pt}
\adjustbox{minipage=\linewidth}{
    \fbox{
        \begin{tabular}{p{\linewidth}}
       
\textbf{[Question]:} \\
How would Others feel as a result? \\
Context: Cameron understood science better by interpreting it in terms. \\
\textbf{[Candidate Answer]:} smart \textcolor{red}{(False)}

\textbf{[Response]:} \\

Answer: [A] \\
The context provided is vague and does not clearly connect Cameron's understanding of science with how others would feel as a result. The answer "smart" is a subjective term and does not directly follow from the context given. Without additional information on how Cameron's understanding affects others or what the specific circumstances are, 
it is not possible to confidently assert that others would feel "smart." 
Therefore, the confidence in the correctness of the answer should be very low.
        \end{tabular}
    }
}
\caption{GPT-4 responses on the negative sample of the Social IQA dataset.}
\label{tab-gpt4}
\end{table*}
\begin{table*}[htbp]
\centering
\small
\setlength{\fboxrule}{1pt} 
\setlength{\fboxsep}{10pt}
\adjustbox{minipage=\linewidth}{
    \fbox{
        \begin{tabular}{p{\linewidth}}
       
\textbf{[Question]:} \\
Determine whether the dialogue summary is correct.\\Dialogue: \\ Person1: I want to go to china for sight-seeing. What do you think of it, Mum? \\ Person2: Why not? China is a wonderful country.\\Person1: Will you go with me, too?\\Person2: No, I'm afraid not now. I'm too busy.\\Summary: Person1 wants to travel to China, but Person1's mother is not available. \\ 
\textbf{[Candidate Answer:]}
Correct \textcolor{red}{(True)}

\textbf{[Response]:} \\
To determine whether the dialogue summary is correct, let's analyze the dialogue and the summary:\\1. Dialogue Analysis:\\   - Person1 expresses a desire to go to China for sightseeing and asks for Person2's opinion.\\    - Person2 responds positively about China, saying it's a wonderful country.\\   - Person1 asks if Person2 will accompany them.\\    - Person2 declines, citing being too busy as the reason.\\ 2. Summary Analysis:\\    - The summary states that Person1 wants to travel to China, which is accurate based on Person1's initial statement.\\    - The summary also states that Person1's mother is not available, which aligns with Person2's response of being too busy to go.\\ 3. Role Identification:\\    - The dialogue does not explicitly identify Person2 as Person1's mother, but it is implied by the use of "Mum" in Person1's question.\\ 4. Conclusion:\\    - The summary accurately captures the main points of the dialogue: Person1's desire to travel to China and Person2's unavailability due to being busy.\\   - The assumption that Person2 is Person1's mother is reasonable given the context.\\ Given this analysis, the answer "Correct" is indeed correct. Therefore, the confidence interval should be high.\\ Answer: [J]
        \end{tabular}
    }
}
\caption{GPT-4o-2024-08-06 responses on the positive sample of the DialogSum dataset.}
\label{tab-gpt4o-2024-08-06}
\end{table*}

\section{Examples of Problems and LLMs Output}\label{Examples of problems and LLMs output}
Taking into account both the openness of the models and their sizes, we provide several examples to demonstrate the effectiveness of our approach. The models vary in size, including those up to 10B, 10-30B, and larger than 30B. Open-source models are presented first, followed by closed-source models. The open-source models are listed in ascending order of size. For further details, please refer to Table \cref{tab-GLM4-chat-9B,tab-Qwen2.5-14B-Instruct,tab-Yi-1.5-34B-Chat-16K,tab-Llama-3.1-70B-Instruct,tab-Qwen-max,tab-gpt4,tab-gpt4o-2024-08-06}. 

\end{document}